\documentclass[headings=standardclasses]{scrartcl}

\usepackage{geometry}
\usepackage[utf8]{inputenc}
\usepackage[T1]{fontenc}
\usepackage[onehalfspacing]{setspace}
\usepackage{charter}
\usepackage[charter]{mathdesign}
\usepackage{microtype}
\usepackage{amsmath}
\usepackage{amsthm}
\usepackage{tabulary}
\usepackage{longtable}
\usepackage{booktabs} 
\usepackage{diagbox}
\usepackage{ragged2e}
\newcommand{\ra}[1]{\renewcommand{\arraystretch}{#1}}
\usepackage{siunitx}
\usepackage{bm}
\usepackage[pdftex]{graphicx}
\usepackage{rotating}
\usepackage{pdfpages}
\usepackage{pdflscape}
\usepackage{color}
\usepackage{array}
\usepackage[font=bf]{caption}
\usepackage{subcaption}
\usepackage{pgfplots}
\usepackage{pgfplotstable}
\usepackage{float}
\usepackage[title]{appendix}
\pgfplotsset{compat=1.9}
\usepackage{url}
\usepackage[section]{placeins}
\usepackage{natbib}
\usepackage{hyperref}
\hypersetup{
	colorlinks   = true,    
	urlcolor     = blue,    
	linkcolor    = blue,    
	citecolor    = blue      
}

\usepackage[boxed]{algorithm2e}
\SetKwFunction{KwInit}{Initialisation}
\SetKwComment{Comment}{\% }{}

\makeatletter
\renewcommand*\env@matrix[1][c]{\hskip -\arraycolsep
  \let\@ifnextchar\new@ifnextchar
  \array{*\c@MaxMatrixCols #1}}
\makeatother

\usepackage{xcolor}
\usepackage{soul}

\usepackage{verbatim}

\begin{document}
\thispagestyle{empty}
\begin{spacing}{1.2}
\begin{flushleft}
\huge \textbf{Bayesian Estimation of Mixed Multinomial Logit Models: Advances and Simulation-Based Evaluations} \\
\vspace{\baselineskip}
\normalsize
12 December 2019 \\
\vspace{\baselineskip}
Prateek Bansal\textsuperscript{*} (corresponding author) \\
School of Civil and Environmental Engineering \\
Cornell University, United States \\
pb422@cornell.edu \\
\vspace{\baselineskip}
Rico Krueger\textsuperscript{*} \\
Research Centre for Integrated Transport Innovation, School of Civil and Environmental Engineering, UNSW Australia, Sydney NSW 2052, Australia \\
r.krueger@student.unsw.edu.au \\
\vspace{\baselineskip}
Michel Bierlaire \\
Transport and Mobility Laboratory, School of Architecture, Civil and Environmental Engineering, Ecole Polytechnique F\'{e}d\'{e}rale de Lausanne, Station 18, Lausanne 1015, Switzerland \\
michel.bierlaire@epfl.ch \\
\vspace{\baselineskip}
Ricardo A. Daziano \\
School of Civil and Environmental Engineering \\
Cornell University, United States  \\
daziano@cornell.edu \\
\vspace{\baselineskip}
Taha H. Rashidi  \\
Research Centre for Integrated Transport Innovation, School of Civil and Environmental Engineering, UNSW Australia, Sydney NSW 2052, Australia\\
rashidi@unsw.edu.au\\
\vspace{\baselineskip}
\textsuperscript{*} These authors contributed equally to this work.
\end{flushleft}
\end{spacing}

\newpage
\thispagestyle{empty}
\section*{Abstract}

Variational Bayes (VB) methods have emerged as a fast and computationally-efficient alternative to Markov chain Monte Carlo (MCMC) methods for scalable Bayesian estimation of mixed multinomial logit (MMNL) models. It has been established that VB is substantially faster than MCMC at practically no compromises in predictive accuracy. In this paper, we address two critical gaps concerning the usage and understanding of VB for MMNL. First, extant VB methods are limited to utility specifications involving only individual-specific taste parameters. Second, the finite-sample properties of VB estimators and the relative performance of VB, MCMC and maximum simulated likelihood estimation (MSLE) are not known. To address the former, this study extends several VB methods for MMNL to admit utility specifications including both fixed and random utility parameters. To address the latter, we conduct an extensive simulation-based evaluation to benchmark the extended VB methods against MCMC and MSLE in terms of estimation times, parameter recovery and predictive accuracy. The results suggest that all VB variants with the exception of the ones relying on an alternative variational lower bound constructed with the help of the modified Jensen’s inequality perform as well as MCMC and MSLE at prediction and parameter recovery. In particular, VB with nonconjugate variational message passing and the delta-method (VB-NCVMP-$\Delta$) is up to 16 times faster than MCMC and MSLE. Thus, VB-NCVMP-$\Delta$ can be an attractive alternative to MCMC and MSLE for fast, scalable and accurate estimation of MMNL models. 
\\
\\
\textit{Keywords:} Variational Bayes; Bayesian inference; mixed logit; nonconjugate variational message passing.


\newpage
\pagenumbering{arabic}

\section{Introduction}

The mixed multinomial logit (MMNL) model \citep{mcfadden2000mixed} is the workhorse model in many disciplines---such as economics, health, marketing and transportation---that are concerned with the analysis and prediction of individual choice behavior. While maximum simulated likelihood estimation \citep[MSLE; see][]{train2009discrete} is the predominant estimation strategy for MMNL models, the Bayesian approach represents an alternative estimation strategy, which entails the key benefit that the whole posterior distribution of all model parameters including the individual-specific parameters can be obtained. Posterior inference in MMNL models is typically performed with the help of Markov chain Monte Carlo (MCMC) methods, which approximate the posterior distribution of the MMNL model parameters through samples from a Markov chain whose stationary distribution is the posterior distribution of interest \citep[see][]{rossi2012bayesian, train2009discrete}. While MCMC methods constitute a powerful framework for posterior inference in complex probabilistic models \citep[see e.g.][]{gelman2013bayesian}, these methods are subject to several bottlenecks, which inhibit their scalability to large datasets, namely i) long computation times, ii) high costs for the storage of the posterior draws and iii) difficulties in assessing convergence \citep{blei2017variational, braun2010variational, depraetere2017comparison, tan2017stochastic}. 

Variational Bayes (VB) methods \citep[e.g.][]{blei2017variational, jordan1999introduction, ormerod2010explaining} have emerged as an alternative to MCMC and promise to address the shortcomings of MCMC methods. The basic intuition behind VB is to view approximate Bayesian inference as an optimization problem rather than a sampling problem. VB aims at finding a parametric variational distribution over the unknown model parameters, whereby the parameters of the variational distribution are optimized such that the probability distance (typically measured in terms of the Kullback-Leibler divergence) between the exact posterior distribution and the variational distribution is minimal. A key challenge in the application of VB to posterior inference in MMNL models is that the expectation of the logarithm of the choice probabilities---or, to be precise, the expectation of the log-sum of exponentials (E-LSE) term---cannot be expressed in closed form, because there is no general conjugate prior for the multinomial logit model. As a consequence, updates for variational factors pertaining to utility parameters require special treatment. The literature proposes different methods to facilitate VB for posterior inference in MMNL models \citep{braun2010variational, depraetere2017comparison, tan2017stochastic}. In essence, these approaches proceed as follows: The E-LSE term is approximated either analytically or by simulation, or an alternative variational lower bound is defined. Then, updates for the nonconjugate variational factors are performed with the help of either quasi-Newton (QN) methods \citep[e.g.][]{nocedal2006numerical} or the nonconjugate variational message passing (NCVMP) approach \citep{knowles2011non}.

Extant studies of VB methods for posterior inference in MMNL models establish that VB is substantially faster than MCMC at negligible compromises in predictive accuracy \citep{braun2010variational, depraetere2017comparison, tan2017stochastic}. However, these studies find wanting in several important ways. First, the QN and NCVMP updating strategies have been studied in isolation from each other and their relative performance is not known. Second, none of these studies compare VB to the widely-used MSLE method. Third, the performance of the considered estimation approaches has only been evaluated in terms of predictive accuracy, while the finite sample properties, i.e. the ability to recover true parameters, of the estimators are not known. Fourth, VB methods have only been implemented and tested for posterior inference in MMNL models with only individual-specific utility parameters despite the practical relevance of fixed utility parameters in discrete choice modeling applications. 

Consequently, the objective of this paper is twofold: First, we extend several VB methods to allow for posterior inference in MMNL models with a more general utility specification including both fixed and random utility parameters.\footnote{Strictly, all model parameters are random quantities in Bayesian estimation. Here, we adopt the nomenclature used by \citet{train2009discrete} and refer to utility parameters that are invariant across decision-makers as fixed utility parameters and to utility parameters that are individual-specific and (normally) distributed across decision-makers as random utility parameters.} Then, we carry out a comprehensive simulation-based evaluation, in which we contrast the performance of different VB methods, MCMC and MSLE in terms of estimation times, parameter recovery and predictive accuracy.\footnote{In this paper, we compare VB and MCMC with MSLE, as MSLE continues to represent the most widely used estimation strategy for MMNL models. We acknowledge that Bhat and co-authors have developed the Maximum Approximate Composite Marginal Likelihood  (MACML) approach \citep{bhat2011simulation} for frequentist estimation of mixed multinomial probit (MMNP) models. MACML has been shown to be faster and more accurate than MSLE \citep{patil2017simulation}. In addition, the approach is flexible, as it has been used for the estimation of integrated choice and latent variable models \citep{bhat2014new} and MMNP models with non-normal parametric mixing distributions \citep{bhat2018new}. Despite its limitation to MMNP, MACML thus represents an attractive alternative to MSLE for frequentist estimation of mixed random utility models. However, we concur with \citet{bhat2018new} that MMNP is no more or less general than MMNL. Comparisons between Bayesian estimation methods for MMNL and MACML for MMNP are admittedly intriguing but are beyond the scope of the current paper.}

We emphasize that the inclusion of fixed utility parameters, in addition to individual-specific utility parameters, is important in practice \citep{bansal2018extending}: First, alternative-specific fixed effects can be introduced by including alternative-specific constants (ASCs) in the utility specification. Assuming ASCs to be random may result in empirical identification issues, especially if their distribution is similar to that of the error term \citep{train2009discrete}. Second, utility parameters corresponding to individual-specific characteristics (e.g. age, gender etc.) are typically assumed to be fixed. Treating these alternative-specific parameters as random may not provide substantive behavioral insights and may unnecessarily inflate the number of random parameters so that the ``curse of dimensionality'' becomes a concern \citep[also see][]{cherchi2012monte}. Third, systematic taste variation can be parsimoniously represented through the inclusion of additional fixed parameters that pertain to interactions of the alternative-specific attribute (e.g. cost or travel time) and relevant individual-specific attributes \citep[e.g. age, household income etc.; see][]{bhat1998accommodating}.

In the case of MSLE, one can easily accommodate fixed utility parameters by specifying them as random utility parameters with a constrained variance, because the individual-specific parameters are integrated out so that that the fixed parameters can be jointly updated with the parameters of the mixing distribution. This approach is not feasible for Bayesian estimation methods, because the individual-specific parameters are directly estimated \citep[see][for the MCMC sampler]{train2009discrete, rossi2012bayesian}. If the fixed utility parameters were specified as random with a constrained variance in VB estimation, the respective variational factors would have to be identical across decision-makers. However, it is not straightforward to impose this restriction in the existing VB methods. This is because the variational factors of the individual-specific parameters are updated independently for each individual, while updates for the variational factors of the fixed parameters necessarily depend on all observations. 

We organize the remainder of this paper as follows: First, we provide a fully Bayesian formulation of the MMNL model (Section \ref{sec:MMNL}). To be self-contained, we present the default MCMC method for posterior inference in MMNL models (Section \ref{subsec:AT}). Then, we describe different VB methods for posterior inference in MMNL models with a more general utility specification including a combination of both fixed and individual-specific utility parameters (Section \ref{S:vb}). Next, we present the simulation-based evaluation (Section \ref{sec:simeval}) and finally, we conclude (Section \ref{sec:concl}). 

\section{Mixed multinomial logit model} \label{sec:MMNL}

The mixed multinomial logit (MMNL) model \citep{mcfadden2000mixed} is established as follows: We consider a standard discrete choice setup, in which on choice occasion $t \in \{1, \ldots T_{n} \}$, a decision-maker $n \in \{1, \ldots N \}$ derives utility $U_{ntj} = V(\boldsymbol{X}_{ntj}, \boldsymbol{\Gamma}_{n}) + \epsilon_{ntj}$ from alternative $j$ in the set $C_{nt}$. Here, $V()$ denotes the representative utility, $\boldsymbol{X}_{ntj}$ is a row-vector of covariates, $\boldsymbol{\Gamma}_{n}$ is a collection of taste parameters, and $\epsilon_{ntj}$ is a stochastic disturbance. The assumption $\epsilon_{ntj} \sim \text{Gumbel}(0,1)$ leads to a multinomial logit (MNL) kernel such that the probability that decision-maker $n$ chooses alternative $j \in C_{nt}$ on choice occasion $t$ is 
\begin{equation}
P(y_{nt} = j \vert \boldsymbol{X}_{ntj}, \boldsymbol{\Gamma}_{n}) = \frac{\exp \left \{ V (\boldsymbol{X}_{ntj}, \boldsymbol{\Gamma}_{n}) \right \}}{\sum_{k \in C_{nt}}\exp \left \{ V (\boldsymbol{X}_{ntk}, \boldsymbol{\Gamma}_{n}) \right \}},
\end{equation} 
where $y_{nt} \in C_{nt}$ captures the observed choice. The choice probability can be iterated over choice scenarios to obtain the probability of observing a decision-maker's sequence of choices $\boldsymbol{y}_{n}$:
\begin{equation}
P(\boldsymbol{y}_{n} \vert \boldsymbol{X}_{n},  \boldsymbol{\Gamma}_{n}) = \prod_{t = 1}^{T_{n}} P(y_{nt} = j \vert \boldsymbol{X}_{nt},  \boldsymbol{\Gamma}_{n}).
\end{equation}

In this paper, we consider a general utility specification under which tastes $\boldsymbol{\Gamma}_{n}$ are partitioned into fixed taste parameters $\boldsymbol{\alpha}$, which are invariant across decision-makers, and random taste parameters $\boldsymbol{\beta}_{n}$, which are individual-specific, such that $\boldsymbol{\Gamma}_{n} = \begin{bmatrix} \boldsymbol{\alpha}^{\top} & \boldsymbol{\beta}_{n}^{\top} \end{bmatrix} ^{\top}$, whereby $\boldsymbol{\alpha}$ and $\boldsymbol{\beta}_{n}$ are vectors of lengths $L$ and $K$, respectively. Analogously, the row-vector of covariates $\boldsymbol{X}_{ntj}$ is partitioned into attributes $\boldsymbol{X}_{ntj,F}$, which pertain to the fixed parameters $\boldsymbol{\alpha}$, as well as into attributes $\boldsymbol{X}_{ntj,R}$, which pertain to the individual-specific parameters $\boldsymbol{\beta}_{n}$, such that $\boldsymbol{X}_{ntj} = \begin{bmatrix} \boldsymbol{X}_{ntj,F} & \boldsymbol{X}_{ntj,R} \end{bmatrix}$. For simplicity, we assume that the representative utility is linear-in-parameters, i.e. 
\begin{equation}
V (\boldsymbol{X}_{ntj}, \boldsymbol{\Gamma}_{n}) = \boldsymbol{X}_{ntj} \boldsymbol{\Gamma}_{n} = \boldsymbol{X}_{ntj,F} \boldsymbol{\alpha} + \boldsymbol{X}_{ntj,R} \boldsymbol{\beta}_{n}.
\end{equation}

The distribution of tastes $\boldsymbol{\beta}_{1:N}$ is assumed to be multivariate normal, i.e. $\boldsymbol{\beta}_{n} \sim \text{N}(\boldsymbol{\zeta}, \boldsymbol{\Omega})$ for $n = 1, \dots, N$, where $\boldsymbol{\zeta}$ is a mean vector and $\boldsymbol{\Omega}$ is a covariance matrix. In a fully Bayesian setup, the invariant (across individuals) parameters $\boldsymbol{\alpha}$, $\boldsymbol{\zeta}$, $\boldsymbol{\Omega}$ are also considered to be random parameters and are thus given priors. We use normal priors for the fixed parameters $\boldsymbol{\alpha}$ and for the mean vector $\boldsymbol{\zeta}$. Following \citet{tan2017stochastic} and \citet{akinc2018Bayesian}, we employ Huang's half-t prior \citep{huang2013Simple} for covariance matrix $\boldsymbol{\Omega}$, as this prior specification exhibits superior noninformativity properties compared to other prior specifications for covariance matrices \citep{huang2013Simple,akinc2018Bayesian}. In particular, \citep{akinc2018Bayesian} show that Huang's half-t prior outperforms the inverse Wishart prior, which is often employed in fully Bayesian specifications of MMNL models \citep[e.g.][]{train2009discrete}, in terms of parameter recovery.

Stated succinctly, the generative process of the fully Bayesian MMNL model is:
\begin{align}
& \boldsymbol{\alpha} \lvert \boldsymbol{\lambda}_{0},\boldsymbol{\Xi}_{0} \sim \text{N}(\boldsymbol{\lambda}_{0},\boldsymbol{\Xi}_{0})\\
& \boldsymbol{\zeta} \vert \boldsymbol{\mu}_{0},\boldsymbol{\Sigma}_{0} \sim \text{N}(\boldsymbol{\mu}_{0},\boldsymbol{\Sigma}_{0}) \\
& a_{k} \lvert A_{k} \sim \text{Gamma}\left( \frac{1}{2}, \frac{1}{A_{k}^{2}} \right), & & k = 1,\dots,K,  \label{eq_gamma_a} \\
& \boldsymbol{\Omega} \vert \nu, \boldsymbol{a}\sim \text{IW}\left(\nu + K - 1, 2\nu \text{diag}(\boldsymbol{a}) \right),  \quad \boldsymbol{a} = \begin{bmatrix} a_{1} & \dots & a_{K} \end{bmatrix}^{\top} \label{eq_iv_Omega} \\
& \boldsymbol{\beta}_{n} \vert \boldsymbol{\zeta}, \boldsymbol{\Omega} \sim \text{N}(\boldsymbol{\zeta}, \boldsymbol{\Omega}), & & n = 1,\dots,N, \\
& y_{nt} \vert \boldsymbol{\alpha}, \boldsymbol{\beta}_{n}, \boldsymbol{X}_{nt} \sim \text{MNL}(\boldsymbol{\alpha}, \boldsymbol{\beta}_{n}, \boldsymbol{X}_{nt}), & & n = 1,\dots,N,  \ t = 1,\dots,T_{n},
\end{align} 
where (\ref{eq_gamma_a}) and (\ref{eq_iv_Omega})  induce Huang's half-t prior \citep{huang2013Simple}. $\{ \boldsymbol{\lambda}_{0}, \boldsymbol{\Xi}_{0}, \boldsymbol{\mu}_{0}, \boldsymbol{\Sigma}_{0}, \nu, A_{1:K} \}$ are known hyper-parameters, and $\boldsymbol{\theta} = \{ \boldsymbol{\alpha}, \boldsymbol{\zeta}, \boldsymbol{\Omega},  \boldsymbol{a}, \boldsymbol{\beta}_{1:N}\}$ is a collection of model parameters whose posterior distribution we wish to estimate. 

The generative process implies the following joint distribution of data and model parameters:
\begin{equation}
\begin{split}
P (\boldsymbol{y}_{1:N}, \boldsymbol{\theta}) = &
\left ( \prod_{n=1}^{N}  P(\boldsymbol{y}_{n} \vert \boldsymbol{X}_{n}, \boldsymbol{\Gamma}_{n}) \right )
P(\boldsymbol{\alpha} \vert \boldsymbol{\lambda}_{0},\boldsymbol{\Xi}_{0})
\left ( \prod_{n=1}^{N} P(\boldsymbol{\beta}_{n} \vert \boldsymbol{\zeta}, \boldsymbol{\Omega}) \right ) \\
& P(\boldsymbol{\zeta} \vert \boldsymbol{\mu}_{0},\boldsymbol{\Sigma}_{0})
P(\boldsymbol{\Omega} \vert \omega, \boldsymbol{B})
\left ( \prod_{k=1}^{K} P(a_{k} \lvert  s,  r_{k}) \right ),
\end{split} 
\end{equation}
where 
$\omega = \nu + K - 1$, 
$\boldsymbol{B} = 2\nu \text{diag}(\boldsymbol{a})$, 
$s = \frac{1}{2}$ and
$r_{k} = A_{k}^{-2}$.\footnote{To be clear, the following forms of the Gamma and inverse Wishart distributions are considered:
\begin{align}
P(a_{k} \vert s,  r_{k})  & \propto a_{k}^{s - 1} \exp(-r_{k} a_{k}), \nonumber \\
P(\boldsymbol{\Omega} \vert \omega, \boldsymbol{B}) & \propto
\vert \boldsymbol{B} \vert ^{\frac{\omega}{2}}
\vert \boldsymbol{\Omega} \vert ^{-\frac{\omega + K + 1}{2}}
\exp \left ( -\frac{1}{2} \text{tr} \left (\boldsymbol{B} \boldsymbol{\Omega}^{-1} \right ) \right ), \nonumber
\end{align}
whereby $\boldsymbol{\Omega}$ and $\boldsymbol{B}$ are $K \times K$ positive-definite matrices.} By Bayes' rule, the posterior distribution of interest is then given by
\begin{equation} \label{eq:post}
P(\boldsymbol{\theta} \vert \boldsymbol{y}_{1:N}) 
= \frac{P (\boldsymbol{y}_{1:N}, \boldsymbol{\theta})}{\int P (\boldsymbol{y}_{1:N}, \boldsymbol{\theta}) d \boldsymbol{\theta}}
\propto P (\boldsymbol{y}_{1:N}, \boldsymbol{\theta}).
\end{equation}
Exact inference of this posterior distribution is not possible, because the model evidence $\int P (\boldsymbol{y}_{1:N}, \boldsymbol{\theta}) d \boldsymbol{\theta}$ is not tractable. In the following sections, we discuss different strategies to approximate the posterior distribution of the MMNL model parameters and provide our extensions to some of these strategies under the more general linear-in-parameters utility specification including both fixed and and random taste parameters.  

\section{Markov chain Monte Carlo} \label{subsec:AT}

The general idea of Markov chain Monte Carlo (MCMC) methods is to approximate a difficult-to-compute posterior distribution through samples from a Markov chain whose stationary distribution is the posterior distribution of interest \citep[see][for a general treatment]{robert2004monte}. 

In the present application, a Markov chain for the posterior distribution of the MMNL model parameters $\boldsymbol{\theta}$ can be constructed by taking samples from the conditional distributions of $\boldsymbol{\theta}$. Direct sampling from the conditional distributions of $\boldsymbol{\zeta}$, $\boldsymbol{\Omega}$ and  $\boldsymbol{a}$ is possible, because the conditional distributions belong to known families of distributions. However, updates for $\boldsymbol{\alpha}$ and $\boldsymbol{\beta}_{1:N}$ need to be generated with the help of random-walk (RW) Metropolis algorithms, because the nonconjugacy of the multinomial logit kernel and the normal priors leads to unrecognizable conditional distributions. The resulting MCMC algorithm is a blocked Gibbs sampler with two embedded Metropolis steps. A pseudo-code representation of the sampler is shown in Algorithm \ref{alg:MH}. Here, $\rho_{\boldsymbol{\alpha}}$ and $\rho_{\boldsymbol{\beta}}$ denote step sizes, which need to be tuned.\footnote{In the subsequent applications of the sampling scheme, we apply the same tuning mechanism as \citet{train2009discrete}, i.e. we let $\rho_{\boldsymbol{\alpha}} = 0.01$ and set $\rho_{\boldsymbol{\beta}}$ to an initial value of 0.1. After each iteration, $\rho_{\boldsymbol{\beta}}$ is decreased by 0.001, if the average acceptance rate across all decision-makers is less than 0.3; $\rho_{\boldsymbol{\beta}}$ is increased by 0.001, if the average acceptance rate across all decision-makers is more than 0.3.} The sampling scheme outlined in Algorithm \ref{alg:MH} is identical to the one studied by \citet{akinc2018Bayesian} with the only difference that updates for the fixed parameters $\boldsymbol{\alpha}$ are incorporated. It is also known as the Allenby-Train procedure \citep{rossi2012bayesian, train2009discrete}. 

A bottleneck of Algorithm \ref{alg:MH} is its reliance on two RW Metropolis steps for the fixed and the individual-specific parameters, respectively. Notwithstanding that these steps are easy to implement and to vectorize, the RW Metropolis algorithm can be inefficient when it is tuned suboptimally \citep[see e.g.][]{rossi2012bayesian}. If the step size is too small, the chain moves too quickly and the draws exhibit high serial correlation. If the step size is too large, the posterior is not properly explored and the algorithm can get stuck. The RW Metropolis algorithm can be replaced by an independence Metropolis algorithm \citep{rossi2012bayesian}, which takes draws around the posterior mode. However, a complication of this approach is that at each iteration, a maximization needs to be performed to find the posterior mode, which is particularly challenging for the individual-specific parameters.

An emerging method to generate samples from a Markov chain is Hamiltonian Monte Carlo \citep[HMC; e.g.][]{neal2011mcmc}. HMC uses information contained in the gradient of the log target density to efficiently explore the posterior distribution of interest and to reduce the amount of serial correlation in the chains. A variant of HMC is the No-U-Turn sampler \citep{hoffman2014no-u-turn}, which automatically adapts the number of leapfrog steps required for the discretization of the Hamiltonian dynamics underlying HMC. NUTS is interfaced by Stan \citep{carpenter2017stan}, a probabilistic programming language that enables posterior inference on a wide variety of user-defined models. However, the generality of Stan comes at an immense computational cost, which is further aggravated when the model of interest depends on many parameters as is the case for MMNL.\footnote{
We also explored the use of Stan as part of the current research study but found that estimation times were prohibitive for the sample sizes considered in the simulation evaluation presented in Section \ref{sec:simeval}. Our experiences with Stan are generally consistent with the literature. \citet{ben2019foundations} contrast NUTS with the Allenby-Train procedure and find that both methods perform equally well at recovering the true parameter values. However, whereas the reported estimation time for the Allenby-Train procedure is 12 minutes, NUTS had to be run ``overnight''. \citet{vij2017random} attempted to use Stan to estimate a MMNL model on a large dataset containing 30,166 observations from 17,700 individuals but were unable to do so due to memory constraints. A possible avenue for future research is to custom-code a NUTS procedure with analytical gradients to enable fast and scalable posterior inference for MMNL.}

\begin{algorithm}[h]
\SetAlgoLined
        \For{1 \KwTo max-iteration}{
          Update $\boldsymbol{\zeta}$ by sampling $\boldsymbol{\zeta} \sim \text{N}\left ( \frac{1}{N} \sum_{n=1}^{N} \boldsymbol{\beta}_{n}, \frac{\boldsymbol{\Omega}}{N} \right )$  \; 
          Update $\boldsymbol{\Omega}$ by sampling $\boldsymbol{\Omega} \sim \text{IW} \left (\nu + N + K - 1, 2 \nu \text{diag}(\boldsymbol{a})  + \sum_{n=1}^{N} (\boldsymbol{\beta}_{n} -  \boldsymbol{\zeta}) (\boldsymbol{\beta}_{n} -  \boldsymbol{\zeta})^{\top} \right )$\;
          Update $a_{k}$ for all $k \in \{1, \ldots, K\}$ by sampling $a_{k} \sim \text{Gamma}\left ( \frac{\nu + K}{2},  \frac{1}{A_{k}^{2}} + \nu \left ( \boldsymbol{\Omega}^{-1} \right )_{kk} \right )$ \;
        
        Update $\boldsymbol{\beta}_{n}$ for all $n \in \{1, \ldots, N\}$:
        \begin{itemize}
        \itemsep0em 
          \item Propose $\tilde{\boldsymbol{\beta}}_{n} = \boldsymbol{\beta}_{n} + \sqrt{\rho_{\boldsymbol{\beta}}} \text{chol}(\boldsymbol{\Omega}) \boldsymbol{\eta}$, where $\boldsymbol{\eta} \sim \text{N}(\boldsymbol{0},\boldsymbol{I}_{K})$ \;
          \item Compute $r = 
            \frac{P(\boldsymbol{y}_{n} \vert \boldsymbol{X}_{n}, \boldsymbol{\alpha}, \tilde{\boldsymbol{\beta}}_{n}) \phi( \tilde{\boldsymbol{\beta}}_{n} \vert \boldsymbol{\zeta},  \boldsymbol{\Omega})}{P(\boldsymbol{y}_{n} \vert \boldsymbol{X}_{n}, \boldsymbol{\alpha}, \boldsymbol{\beta}_{n}) \phi( \boldsymbol{\beta}_{n} \vert \boldsymbol{\zeta},  \boldsymbol{\Omega})}$\;
         \item Draw $u \sim \text{Uniform}(0,1)$. If $r \leq u$, accept the proposal, else reject it.
       \end{itemize}
 
       Update $\boldsymbol{\alpha}$:
         \begin{itemize}
         \itemsep0em
            \item Propose $\tilde{\boldsymbol{\alpha}} = \boldsymbol{\alpha} + \sqrt{\rho_{\boldsymbol{\alpha}}} \text{chol}(\boldsymbol{\Xi}_{0}) \boldsymbol{\eta}$, where $\boldsymbol{\eta} \sim \text{N}(\boldsymbol{0},\boldsymbol{I_{L}})$;
            \item Compute $r = 
            \frac{\prod_{n = 1}^{N} P(\boldsymbol{y}_{n} \vert \boldsymbol{X}_{n}, \tilde{\boldsymbol{\alpha}}, \boldsymbol{\beta}_{n}) \phi( \tilde{\boldsymbol{\alpha}} \vert \boldsymbol{\lambda}_{0},  \boldsymbol{\Xi}_{0})}
            {\prod_{n = 1}^{N} P(\boldsymbol{y}_{n} \vert \boldsymbol{X}_{n}, \boldsymbol{\alpha}, \boldsymbol{\beta}_{n}) \phi( \boldsymbol{\alpha}  \vert \boldsymbol{\lambda}_{0},  \boldsymbol{\Xi}_{0})}$;
            \item Draw $u \sim \text{Uniform}(0,1)$. If $r \leq u$, accept the proposal, else reject it. 
         \end{itemize}
		 }
\caption{Pseudo-code representation of the blocked Gibbs sampler for posterior inference in MMNL models with fixed and random utility parameters} \label{alg:MH}
\end{algorithm}

\section{Variational Bayes} \label{S:vb}

\subsection{Background}

\begin{figure}[H]
\centering
\includegraphics[width = \textwidth]{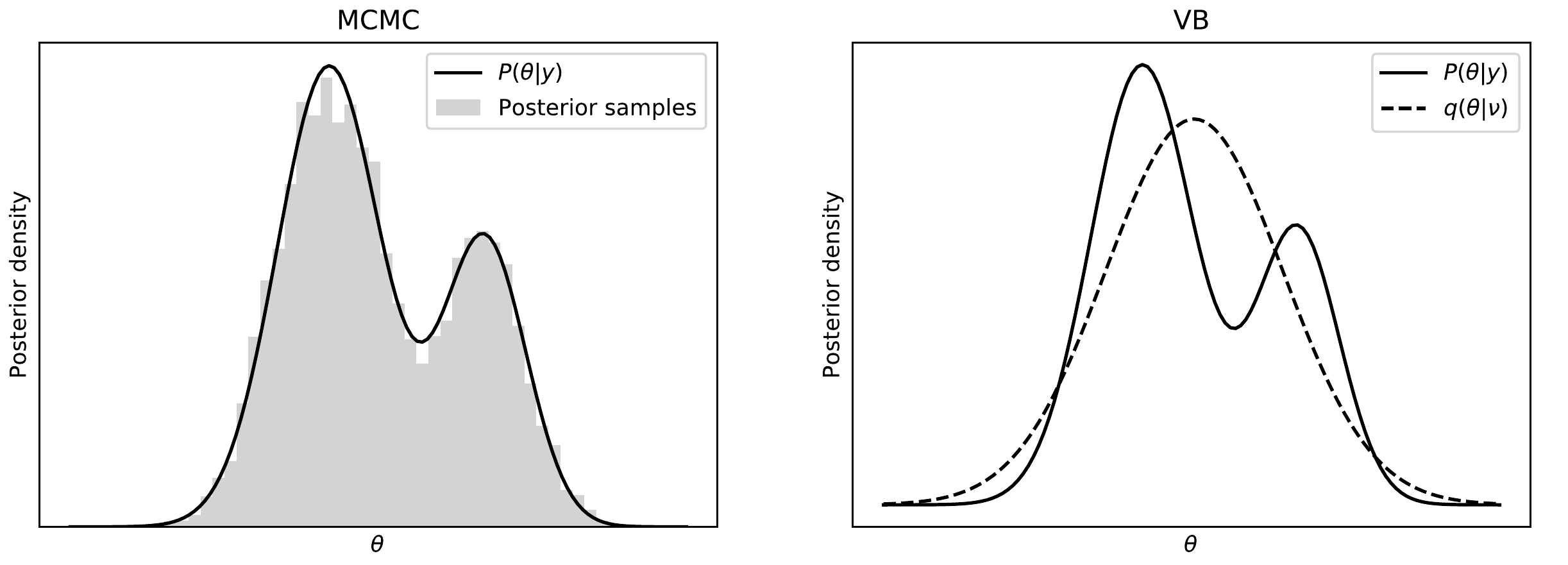}
\caption[Schematic representations of Markov chain Monte Carlo and Variational Bayes methods for posterior inference]{Schematic representations of Markov chain Monte Carlo (MCMC) and Variational Bayes (VB) methods for posterior inference} \label{f_vb}
\end{figure}

Variational Bayes \citep[VB; e.g.][]{blei2017variational, jordan1999introduction, ormerod2010explaining} differs from MCMC in that approximate Bayesian inference is viewed as optimization problem rather than a sampling problem. Figure \ref{f_vb} illustrates the conceptual differences between MCMC and VB. In MCMC, the posterior distribution of interest $P(\theta \vert \boldsymbol{y})$ is approximated through samples from a Markov chain whose stationary distribution is the posterior distribution of interest. In VB, the posterior distribution of interest is approximated through a parametric variational distribution $q(\theta \vert \boldsymbol{\nu})$ whose parameters $\boldsymbol{\nu}$ are fit such that the $P(\theta \vert \boldsymbol{y})$ and the approximating variational distribution are close in probability distance.

Casting approximate Bayesian inference as an optimization problem comes with several benefits which enable scaling Bayesian estimation to large datasets. First, the memory issues of MCMC are overcome, as only the variational parameters rather than the posterior draws need to be stored. Second, convergence can be straightforwardly assessed by evaluating the change in the variational lower bound (an alternative measure for the distance between the posterior distribution of interest and the approximating variational distribution) or the change in the estimates of the variational parameters from one iteration to another. Third, serial correlation is no longer a concern, as no samples are taken.

To build further intuition about the fundamental principles of VB, we consider a generative model $P(\boldsymbol{y}, \boldsymbol{\theta})$ consisting of observed data $\boldsymbol{y}$ and unknown parameters $\boldsymbol{\theta}$. Our goal is to find an approximation of the posterior distribution $P(\boldsymbol{\theta} \vert \boldsymbol{y})$. VB aims at finding a variational distribution $q(\boldsymbol{\theta})$ over the unknown parameters that is close to the actual posterior distribution $P(\boldsymbol{\theta} \vert \boldsymbol{y})$. A computationally-convenient way to measure the distance between two probability distributions is the Kullback-Leibler (KL) divergence \citep{kullback1951information}. The KL divergence between $q(\boldsymbol{\theta})$ and $P(\boldsymbol{\theta} \vert \boldsymbol{y})$ is given by 
\begin{equation} 
\begin{split} \label{eq:KL}
\text{KL} \left (q(\boldsymbol{\theta}) \vert \vert P(\boldsymbol{\theta} \vert \boldsymbol{y}) \right ) 
& = \int \ln \left ( \frac{q(\boldsymbol{\theta})}{P(\boldsymbol{\theta} \vert \boldsymbol{y})} \right ) q(\boldsymbol{\theta}) d q(\boldsymbol{\theta}) \\
& = \mathbb{E}_{q} \left \{ \ln q(\boldsymbol{\theta}) \right \} - \mathbb{E}_{q} \left \{ \ln P(\boldsymbol{\theta} \vert \boldsymbol{y}) \right \}.
\end{split}
\end{equation}
The goal of VB is to minimize this divergence, i.e.
\begin{equation} 
q^{*}(\boldsymbol{\theta})
= \operatorname*{arg\,min}_{q} \left \{ \text{KL} \left (q(\boldsymbol{\theta}) \vert \vert P(\boldsymbol{\theta} \vert \boldsymbol{y}) \right ) \right \}.
\end{equation}
However, the expectation $\mathbb{E}_{q} \left \{ \ln P(\boldsymbol{\theta} \vert \boldsymbol{y}) \right \} = \mathbb{E}_{q} \left \{ \ln P(\boldsymbol{y}, \boldsymbol{\theta}) \right \} - \ln P(\boldsymbol{y})$ in expression \ref{eq:KL} is not analytically tractable, because there is not closed-form expression for $\ln P(\boldsymbol{y})$. Therefore, we consider the following alternative objective function:
\begin{equation}
\begin{split} \label{eq:ELBO}
\text{KL} \left (q(\boldsymbol{\theta}) \vert \vert P(\boldsymbol{y}, \boldsymbol{\theta}) \right )
& = \text{KL} \left (q(\boldsymbol{\theta}) \vert \vert P(\boldsymbol{\theta} \vert \boldsymbol{y}) \right ) - \ln P(\boldsymbol{y}) \\
& = \mathbb{E}_{q} \left \{ \ln q(\boldsymbol{\theta}) \right \} - \mathbb{E}_{q} \left \{ \ln P(\boldsymbol{y}, \boldsymbol{\theta}) \right \}
\end{split}
\end{equation}
The term $\mathbb{E}_{q} \left \{ \ln P(\boldsymbol{y}, \boldsymbol{\theta}) \right \}- \mathbb{E}_{q} \left \{ \ln q(\boldsymbol{\theta}) \right \}$ is referred to as the evidence lower bound (ELBO). Maximizing the ELBO is equivalent to minimizing the KL divergence between the approximate variational distribution and the intractable exact posterior distribution. Consequently, the goal of VB can be re-formulated as
\begin{equation} 
\begin{split}
q^{*}(\boldsymbol{\theta})
& = \operatorname*{arg\,max}_{q} \left \{ \text{ELBO}(q) \right \} \\
& = \operatorname*{arg\,max}_{q} \left \{ \mathbb{E}_{q} \left \{ \ln P(\boldsymbol{y}, \boldsymbol{\theta}) \right \} - \mathbb{E}_{q} \left \{ \ln q(\boldsymbol{\theta}) \right \} \right \}.
\end{split}
\end{equation}

The functional form of the variational distribution $q(\boldsymbol{\theta})$ remains to be chosen. In principle, the complexity of the variational distribution determines the quality of the approximation of the posterior and the difficulty of the optimisation problem \citep{blei2017variational}. Here, we appeal to the mean-field family of distributions \citep[e.g.][]{jordan1999introduction}, under which the variational distribution factorises as 
\begin{equation}
q(\boldsymbol{\theta}) = \prod_{j=1}^{J} q(\boldsymbol{\theta}_{j}),
\end{equation}
where $j \in \{1, \ldots, J\}$ indexes the model parameters collected in $\boldsymbol{\theta}$. The mean-field assumption breaks the dependence between the model parameters by imposing mutual independence of the variational factors. It can be shown that the optimal density of each variational factor is given by 
\begin{equation}
q^{*}(\boldsymbol{\theta}_{j}) \propto \exp \mathbb{E}_{- \boldsymbol{\theta}_{j}} \left \{ \ln P(\boldsymbol{y}, \boldsymbol{\theta}) \right \},
\end{equation}
i.e. the optimal density of each variational factor is proportional to the exponentiated expectation of the logarithm of the joint distribution of $\boldsymbol{y}$ and $\boldsymbol{\theta}$, where the expectation is taken with respect to all parameters other than $\boldsymbol{\theta}_{j}$ \citep{ormerod2010explaining,blei2017variational}. Provided that the model of interest is conditionally conjugate, the optimal densities of all variational factors belong to recognizable families of distributions \citep{blei2017variational}. Due to the implicit nature of the expectation operator $\mathbb{E}_{- \boldsymbol{\theta}_{j}}$, the ELBO can then be maximized via a simple iterative coordinate ascent algorithm \citep{bishop2006pattern}, in which the variational factors are updated one at a time conditional on the current estimates of the other variational factors. With this algorithm, iterative updates with respect to each variational factor are performed by equating each of the variational factors to its respective optimal density, i.e. we set $q(\boldsymbol{\theta}_{j}) = q^{*}(\boldsymbol{\theta}_{j})$ for $j = 1, \ldots, J$. Because the ELBO is convex with respect to each of the variational factors, the ELBO is guaranteed to converge to a local optimum \citep{boyd2004convex}. Moreover, an important result from the frequentist perspective is the variational Bernstein-von Mises theorem, which states that under benign conditions, the mean-field variational Bayes estimate $\check{\boldsymbol{\theta}} = \int \boldsymbol{\theta} q^{*}(\boldsymbol{\theta}) d \boldsymbol{\theta}$ is consistent \citep{wang2018frequentist}.

Finally, we observe that VB can be viewed as a tractable approximation of the expectation-maximization (EM) algorithm \citep{dempster1977maximum}. To make this analogy clear, we partition the model parameters into global parameters $\boldsymbol{\theta}_{G} = \{ \boldsymbol{\alpha}, \boldsymbol{\zeta}, \boldsymbol{\Omega}, \boldsymbol{a}\}$ and local parameters (latent variables) $\boldsymbol{\theta}_{L} = \boldsymbol{\beta}_{1:N}$. Since the EM algorithm is a frequentist estimation procedure, point estimates (instead of the posterior distribution) of the global parameters $\boldsymbol{\theta}_{G}$ are of interest and are obtained by maximizing the log-likelihood via a two-step iterative procedure. In the expectation step (E-step), the distribution of local parameters conditional on the current estimates of the global parameters is calculated. In the maximization step (M-step), the conditional expectation (i.e. the lower bound on the log-likelihood) is maximized over the unknown global parameters. In Bayesian estimation, the global parameters are also treated as random variables and the posterior distribution of both the local and the global parameters is estimated. VB becomes useful when the conditional expectation relative to these parameters is intractable. Whereas the EM algorithm works with the exact conditional distribution on the local parameters, VB approximates the intractable conditional distributions of the parameters of interest with the help of a simpler, parametric variational distribution. In a similar way as the EM algorithm, VB updates the parameters of the variational distribution by iteratively maximizing the ELBO (which is analogous to the lower bound of the log-likelihood in EM); each VB iteration tightens the gap between the variational distribution and the actual posterior distribution. For more details on the connection between VB and the EM algorithm, we refer to \citet{beal2003variational}.

\subsection{Variational Bayes for posterior inference in mixed multinomial logit models} \label{subsec:vb_mmnl}

\subsubsection{General strategy}

In the present application, we are interested in approximating the posterior distribution of the MMNL model parameters $\{\boldsymbol{\alpha}, \boldsymbol{\zeta}, \boldsymbol{\Omega},  a_{1:K}, \boldsymbol{\beta}_{1:N} \}$ (see expression \ref{eq:post}) through a fitted variational distribution. We posit a variational distribution from the mean-field family, i.e. the variational distribution factorizes as follows:
\begin{equation} 
q(\boldsymbol{\theta})=q(\boldsymbol{\alpha}, \boldsymbol{\zeta}, \boldsymbol{\Omega},  a_{1:K}, \boldsymbol{\beta}_{1:N}) =
q(\boldsymbol{\alpha})
q(\boldsymbol{\zeta})
q(\boldsymbol{\Omega})
\prod_{k=1}^{K} q(a_{k})
\prod_{n=1}^{N} q(\boldsymbol{\beta}_{n}).
\end{equation}
Recall that the optimal densities of the variational factors are given by $q^{*}(\theta_{i}) \propto \exp \mathbb{E}_{- \theta_{i}} \left \{ \ln P(\boldsymbol{y}, \boldsymbol{\theta}) \right \}$.  We find that $q^{*}(\boldsymbol{\zeta} \lvert \boldsymbol{\mu}_{\boldsymbol_{\zeta}}, \boldsymbol{\Sigma}_{\boldsymbol{\zeta}})$, $q^{*}(\boldsymbol{\Omega} \lvert w, \boldsymbol{\Theta})$ and $q^{*}(a_{k} \lvert c, d_k)$ are common probability distributions (see Appendix \ref{A:optQ}). However, $q^{*}(\boldsymbol{\alpha})$ and $q^{*}(\boldsymbol{\beta}_{n})$ are not members of recognizable families of distributions, because the MNL kernel does not have a general conjugate prior. For simplicity and computational convenience, we assume that $q(\boldsymbol{\alpha}) = \text{Normal}(\boldsymbol{\mu}_{\boldsymbol{\alpha}}, \boldsymbol{\Sigma}_{\boldsymbol{\alpha}})$ and $q(\boldsymbol{\beta}_{n}) = \text{Normal}(\boldsymbol{\mu}_{\boldsymbol{\beta}_{n}}, \boldsymbol{\Sigma}_{\boldsymbol{\beta}_{n}})$ for all $n \in \{1, \ldots, N \}$. For notational convenience, we can combine the variational factors such that $q(\boldsymbol{\alpha}) q(\boldsymbol{\beta}_{n})  = q(\boldsymbol{\Gamma}_{n}) = \text{Normal}( \boldsymbol{\Gamma}_{n0}, \boldsymbol{V}_{\boldsymbol{\Gamma}_{n0}})$ with $\boldsymbol{\Gamma}_{n0} = \begin{bmatrix} \boldsymbol{\mu}_{\boldsymbol{\alpha}}^{\top} & \boldsymbol{\mu}_{\boldsymbol{\beta}_{n}}^{\top} \end{bmatrix}^{\top}$ and $\boldsymbol{V}_{\boldsymbol{\Gamma}_{n0}} = \begin{bmatrix}\boldsymbol{\Sigma}_{\alpha} & 0 \\ 0 &  \boldsymbol{\Sigma}_{\boldsymbol{\beta}_{n}} \end{bmatrix}$ for $n = 1, \dots, N$. The negative entropy of the variational distribution is given by
\begin{equation}
\mathbb{E} \left \{ \ln q(\boldsymbol{\theta}) \right \} =
- \frac{1}{2} \ln \vert \boldsymbol{\Sigma_{\alpha}} \vert 
- \frac{1}{2} \ln \vert \boldsymbol{\Sigma_{\zeta}} \vert 
- \frac{K + 1}{2}  \ln \vert \boldsymbol{\Theta} \vert 
+ \sum_{k=1}^{K} \ln d_{k} 
- \frac{1}{2} \sum_{n=1}^{N} \ln \vert \boldsymbol{\Sigma}_{\boldsymbol{\beta}_{n}} \vert.
\end{equation}
Moreover, the logarithm of the joint distribution of the data and the unknown model parameters is given by
\begin{equation}
\begin{split}
\ln P & (\boldsymbol{y}_{1:N}, \boldsymbol{\theta}) \\
= 
& \sum_{n=1}^{N} \ln P(\boldsymbol{y}_{n} \vert \boldsymbol{X}_{n},  \boldsymbol{\Gamma}_{n}) +
\ln P(\boldsymbol{\alpha} \vert \boldsymbol{\lambda}_{0},\boldsymbol{\Xi}_{0})
+\ln P(\boldsymbol{\zeta} \vert \boldsymbol{\mu}_{0},\boldsymbol{\Sigma}_{0}) \\
& + \ln P(\boldsymbol{\Omega} \vert \omega, \boldsymbol{B})
+ \sum_{k=1}^{K} \ln P(a_{k} \vert  s,  r_{k})
+ \sum_{n=1}^{N} \ln P(\boldsymbol{\beta}_{n} \vert \boldsymbol{\zeta}, \boldsymbol{\Omega}) \\
= 
& \sum_{n=1}^{N} \ln P(\boldsymbol{y}_{n} \vert \boldsymbol{X}_{n},  \{\boldsymbol{\alpha}, \boldsymbol{\beta}_{n} \})
-\frac{1}{2} (\boldsymbol{\alpha} - \boldsymbol{\lambda}_{0})^{\top} \boldsymbol{\Xi}_{0}^{-1} (\boldsymbol{\alpha} - \boldsymbol{\lambda}_{0})
-\frac{1}{2} (\boldsymbol{\zeta} - \boldsymbol{\mu}_{0})^{\top} \boldsymbol{\Sigma}_{0}^{-1} (\boldsymbol{\zeta} - \boldsymbol{\mu}_{0})\\ 
& + \frac{\omega}{2}\ln \vert \boldsymbol{B}  \vert- \frac{\omega + K + 1}{2} \ln \vert \boldsymbol{\Omega} \vert - \frac{1}{2} \text{tr} \left ( \boldsymbol{B} \boldsymbol{\Omega}^{-1} \right )
+ \sum_{k=1}^{K} \left[(s - 1) \ln a_{k} - r_{k} a_{k} \right]\\
& - \frac{N}{2} \ln \vert \boldsymbol{\Omega} \vert -\frac{1}{2} \sum_{n=1}^{N} (\boldsymbol{\beta}_{n} -  \boldsymbol{\zeta})^{\top} \boldsymbol{\Omega}^{-1} (\boldsymbol{\beta}_{n} -  \boldsymbol{\zeta}). \\
\end{split}
\end{equation}
Taking expectations, we obtain
\begin{equation} \label{eq:EofJoint}
\begin{split}
\mathbb{E} & \left \{ \ln P  (\boldsymbol{y}_{1:N}, \boldsymbol{\theta} \right \} \\
= 
&  \sum_{n=1}^{N}\sum_{t=1}^{T_{n}} \Bigg\{\sum_{k \in C_{nt}} \left[ y_{ntk}(\boldsymbol{X}_{ntk,F} \boldsymbol{\mu}_{\alpha} +  \boldsymbol{X}_{ntk,R} \boldsymbol{\mu}_{\boldsymbol{\beta}_n} ) \right] -
\mathbb{E}_{q} \Bigg( \ln\Bigg[ \sum_{k \in C_{nt}} \exp(\boldsymbol{X}_{ntk} \boldsymbol{\Gamma}_{n} )\Bigg] \Bigg) \Bigg\} \\
& -\frac{1}{2} (\boldsymbol{\mu}_{\boldsymbol{\alpha}} - \boldsymbol{\lambda}_{0})^{\top} \boldsymbol{\Xi}_{0}^{-1} (\boldsymbol{\mu}_{\boldsymbol{\alpha}} - \boldsymbol{\lambda}_{0}) -\frac{1}{2} \text{tr} \left ( \boldsymbol{\Xi}_{0}^{-1} \boldsymbol{\Sigma}_{\boldsymbol{\alpha}}\right ) \\
& -\frac{1}{2} (\boldsymbol{\mu}_{\boldsymbol{\zeta}} - \boldsymbol{\mu}_{0})^{\top} \boldsymbol{\Sigma}_{0}^{-1} (\boldsymbol{\mu}_{\boldsymbol{\zeta}} - \boldsymbol{\mu}_{0})  -\frac{1}{2} \text{tr} \left ( \boldsymbol{\Sigma}_{0}^{-1} \boldsymbol{\Sigma}_{\boldsymbol{\zeta}} \right )\\
& - \frac{\omega}{2} \sum_{k = 1}^{K} \ln d_{k} - \frac{\omega + K + 1}{2} \ln \vert \boldsymbol{\Theta} \vert - \nu w \sum_{k = 1}^{K} \frac{c}{d_{k}} \left ( \boldsymbol{\Theta}^{-1} \right )_{kk}
+ \sum_{k = 1}^{K} \left [ (1-s) \ln d_{k} - r_{k} \frac{c}{d_{k}} \right ] \\
& - \frac{N}{2} \ln \vert \boldsymbol{\Theta} \vert -\frac{w}{2}\sum_{n=1}^{N} \left[(\boldsymbol{\mu}_{\boldsymbol{\beta}_{n}} - \boldsymbol{\mu}_{\boldsymbol{\zeta}})^{\top}\boldsymbol{\Theta}^{-1}(\boldsymbol{\mu}_{\boldsymbol{\beta}_{n}} - \boldsymbol{\mu}_{\boldsymbol{\zeta}}) + \text{tr} \left ( \boldsymbol{\Theta}^{-1} \boldsymbol{\Sigma}_{\boldsymbol{\beta}_{n}} \right ) +  \text{tr} \left ( \boldsymbol{\Theta}^{-1} \boldsymbol{\Sigma}_{\boldsymbol{\zeta}} \right ) \right]. \\
\end{split}
\end{equation}
Hence, the ELBO of MMNL is:
\begin{equation}
 \text{ELBO} = \mathbb{E}  \left \{ \ln P  (\boldsymbol{y}_{1:N}, \boldsymbol{\theta}) \right \} - \mathbb{E} \left \{ \ln q(\boldsymbol{\theta}) \right \}.
\end{equation}

The ELBO is maximized using an iterative coordinate ascent algorithm. Iterative updates of $q(\boldsymbol{\zeta})$, $q(\boldsymbol{\Omega})$, and $q(a_{k})$ are performed by equating each variational factor to its respective optimal distribution $q^{*}(\boldsymbol{\zeta})$, $q^{*}(\boldsymbol{\Omega})$ and $q^{*}(a_{k})$, respectively. However, updates of $q(\boldsymbol{\alpha})$ and $q(\boldsymbol{\beta}_{n})$ require special treatment, because there is no closed-form expression for the expectation of the log-sum of exponentials (LSE) in equation \ref{eq:EofJoint}. To be precise, the LSE term is given by
\begin{equation} \
g_{nt}(\boldsymbol{\Gamma}_{n}) \equiv \ln \sum_{k \in C_{nt}} \exp(\boldsymbol{X}_{ntk} \boldsymbol{\Gamma}_{n} )  = \ln  \sum_{j \in C_{nt}} \exp( \boldsymbol{X}_{ntj,F} \boldsymbol{\alpha} + \boldsymbol{X}_{ntj,R} \boldsymbol{\beta}_{n}),
\end{equation}
and $\mathbb{E}_{q} \left \{ g_{nt}(\boldsymbol{\Gamma}_{n}) \right \}$ (henceforth, E-LSE) is not tractable.

\subsubsection{Approximations, bounds and updating strategies}

The literature proposes different methods for enabling VB for posterior inference in MMNL models with only individual-specific utility parameters (i.e. $\boldsymbol{\Gamma}_{n} = \boldsymbol{\beta}_{n}$) \citep{braun2010variational, depraetere2017comparison, tan2017stochastic}. In essence, these methods proceed as follows: The E-LSE term is approximated either analytically or by simulation, or an alternative variational lower bound is defined. Then, updates for the nonconjugate variational factors are performed with the help of either quasi-Newton (QN) methods \citep[e.g.][]{nocedal2006numerical} or nonconjugate variational message passing \citep[NCVMP;][]{knowles2011non}. 

Table \ref{table_vb_methods} provides an overview of relevant instances of VB methods for posterior inference in MMNL models and classifies these approaches according to their E-LSE approximation method or lower bound and their updating strategy. Table \ref{table_vb_methods} also shows which methods are extended in the current paper to allow for posterior inference in MMNL models with both fixed and random utility parameters. In this study, we consider one analytical approximation method, namely the Delta ($\Delta$) method \citep[e.g.][]{bickel2015mathematical}, one simulation-based approximation method, namely quasi-Monte Carlo (QMC) integration \citep[e.g.][]{dick2010digital}, as well as an alternative variational lower bound of E-LSE defined with the help of the modified Jensen's inequality \citep[MJI;][]{knowles2011non} in combination with QN- and NCVMP-based updates.\footnote{QMC methods are widely used in statistics and related areas to approximate intractable integrals by simulation. For a general treatment of QMC methods, we refer to \citet{dick2010digital}. For in-depth treatments of QMC methods in the context of simulation-assisted estimation of discrete choice models, the reader is directed to \citet{bhat2001quasi}, \citet{sivakumar2005simulation} and \citet{train2009discrete}.}\textsuperscript{,}\footnote{\label{fn:ncvmp_qmc} In this study, we do not consider NCVMP in combination with QMC integration (henceforth, VB-NCVMP-QMC), as the calculations of the gradients of the expectations of the logarithm of the joint distribution involve inversions of large matrices. As a consequence, VB-NCVMP-QMC becomes numerically unstable and positive-definiteness of the updates of the covariance matrices $\boldsymbol{\Sigma}_{\boldsymbol{\alpha}}$ and $\boldsymbol{\Sigma}_{\boldsymbol{\beta}_{n}}$ cannot be guaranteed. The updates of the nonconjugate variational factors in VB-NCVMP-QMC can be made available upon request.} 

We select the analytical and simulation-based E-LSE approximation methods and the alternative variational lower bound as well as the updating strategies for the nonconjugate variational factors based on the findings of earlier studies: \citet{tan2017stochastic} also adopts the stochastic linear regression (SLR) approach \citep{salimans2013fixed} for posterior inference in MMNL models with only individual-specific utility parameters. SLR is a VB variant, which involves stochastic simulations to update the variational distributions in non-conjugate models. In this paper, we do not extend VB-SLR for posterior inference in MMNL model with a more general utility specification involving a combination of fixed and random utility parameters, because it is computationally expensive to condition the iterative and simulation-based updates of one set of parameters on the approximate posterior distribution of the other set of parameters. \citet{tan2017stochastic} further uses Laplace's method to approximate E-LSE and then employs QN methods to update $q(\boldsymbol{\beta}_{n})$ (henceforth, VB-QN-L). However, VB-QN-L is found to provide inferior predictive accuracy in comparison with MCMC, VB-NCVMP-$\Delta$ and VB-SLR. Moreover, \citet{braun2010variational} also consider the original version of Jensen's inequality to define an alternative variational lower bound and then use QN methods to update $q(\boldsymbol{\beta}_{n})$. However, the modified Jensen's inequality proposed by \citet{knowles2011non} provides a tighter lower bound. \citet{depraetere2017comparison} study a variety of other quadratic lower bounds but find that these bounds are outperformed by the modified Jensen's inequality. From Table \ref{table_vb_methods}, it can further be seen that the relative performance the QN- and NCVMP-based updating strategies are not known, as these updating strategies have been studied in isolation from each other. 

\begin{landscape}
\begin{table}[h]
\centering
\ra{1.2}
\begin{tabular}{c | p{0.33\textwidth} | p{0.33\textwidth} | p{0.33\textwidth}}
\diagbox[height = 4cm]{\textbf{Strategies to}  \\ \textbf{update non-conjugate} \\ \textbf{variational factors}}{\textbf{E-LSE approximation /} \\  \textbf{lower bound}}
&
\begin{minipage}[t]{0.33\textwidth} \centering 
Delta ($\Delta$) method 

\small
\vspace{0.5 \baselineskip}
e.g. \citet{bickel2015mathematical}
\end{minipage} 
&
\begin{minipage}[t]{0.33\textwidth} \centering 
Quasi-Monte Carlo (QMC) integration 

\small
\vspace{0.5 \baselineskip}
e.g. \citet{dick2010digital}
\end{minipage} 
&
\begin{minipage}[t]{0.33\textwidth} \centering 
Modified Jensen's inequality (MJI)

\small
\vspace{0.5 \baselineskip}
\citet{knowles2011non}
\end{minipage} \\
\hline
\begin{minipage}[t]{0.33\textwidth} \centering 
\vspace{0.5 \baselineskip}
Quasi-Newton (QN) methods

\vspace{0.5 \baselineskip}
\small
e.g. \citet{nocedal2006numerical}
\vspace{0.5 \baselineskip} 
\end{minipage} 
&
\begin{minipage}[t]{0.33\textwidth} \centering 
\vspace{0.5 \baselineskip}
\textit{VB-QN-$\Delta$}

\vspace{0.5 \baselineskip}
\small
\citet{braun2010variational}; \citet{depraetere2017comparison}; \textit{this paper}
\vspace{0.5 \baselineskip}
\end{minipage} 
&
\begin{minipage}[t]{0.33\textwidth} \centering 
\vspace{0.5 \baselineskip}
\textit{VB-QN-QMC}

\vspace{0.5 \baselineskip}
\small
\citet{depraetere2017comparison}; \textit{this paper}
\vspace{\baselineskip}
\end{minipage} 
&
\begin{minipage}[t]{0.33\textwidth} \centering 
\vspace{0.5 \baselineskip}
\textit{VB-QN-MJI}

\vspace{0.5 \baselineskip}
\small
\citet{depraetere2017comparison}; \textit{this paper}
\vspace{0.5 \baselineskip}
\end{minipage} \\
\hline
\begin{minipage}[t]{0.33\textwidth} \centering 
\vspace{0.5 \baselineskip}
Nonconjugate variational message passing (NCVMP)

\vspace{0.5 \baselineskip}
\small
\citet{knowles2011non}
\vspace{0.5 \baselineskip} 
\end{minipage} 
&
\begin{minipage}[t]{0.33\textwidth} \centering 
\vspace{0.5 \baselineskip}
\textit{VB-NCVMP-$\Delta$}

\vspace{0.5 \baselineskip}
\small
\citet{tan2017stochastic}; \textit{this paper}
\vspace{0.5 \baselineskip}
\end{minipage} 
&
\begin{minipage}[t]{0.33\textwidth} \centering 
\vspace{0.5 \baselineskip}
\textit{VB-NCVMP-QMC}

\vspace{0.5 \baselineskip}
\small
see footnote \ref{fn:ncvmp_qmc}
\vspace{0.5 \baselineskip}
\end{minipage} 
&
\begin{minipage}[t]{0.33\textwidth} \centering 
\vspace{0.5 \baselineskip}
\textit{VB-NCVMP-MJI}

\vspace{0.5 \baselineskip}
\small
\textit{this paper}
\vspace{0.5 \baselineskip}
\end{minipage} \\
\end{tabular}
\footnotesize{Note: All previous studies exclusively consider utility specifications with only random taste parameters. This paper extends relevant methods to admit utility specifications with both fixed and random taste parameters.}
\caption{Overview of variational Bayes methods for posterior inference in mixed multinomial logit models}
\label{table_vb_methods}
\end{table}
\end{landscape}

In what follows, we describe the considered methods to approximate E-LSE and the alternative variational lower bound:
\begin{enumerate}
\item The Delta ($\Delta$) method involves a second-order Taylor series expansion of $g_{nt}(\boldsymbol{\Gamma}_{n})$ around $\boldsymbol{\Gamma}_{n0}$:
\begin{equation}
g_{nt}(\boldsymbol{\Gamma}_{n}) \approx  
g_{nt}(\boldsymbol{\Gamma}_{n0}) + 
\left ( \boldsymbol{\Gamma}_{n} - \boldsymbol{\Gamma}_{n0} \right )^{\top} \left ( \nabla g_{nt}(\boldsymbol{\Gamma}_{n0}) \right )  +  
\frac{1}{2} \left ( \boldsymbol{\Gamma}_{n} - \boldsymbol{\Gamma}_{n0} \right )^{\top} \left ( \nabla^2 g_{nt}(\boldsymbol{\Gamma}_{n0}) \right)  \left ( \boldsymbol{\Gamma}_{n} - \boldsymbol{\Gamma}_{n0} \right ).
\end{equation}
Then,
\begin{equation}\label{eq:delta}
    \begin{split}
     \mathbb{E}_{q} \{g_{nt}(\boldsymbol{\Gamma}_{n})\}  \approx &  g_{nt}(\boldsymbol{\Gamma}_{n0}) + \frac{1}{2} \text{tr} \left( \nabla^2 g_{nt}(\boldsymbol{\Gamma}_{n0})\boldsymbol{V}_{\boldsymbol{\Gamma}_{n0}} \right ) \\
     \approx & g_{nt}(\boldsymbol{\Gamma}_{n0}) + \frac{1}{2} \text{tr}\left ( \frac{\partial^2 g_{nt}(\boldsymbol{\Gamma}_{n0})}{\partial \boldsymbol{\beta}_{n}^2}  \boldsymbol{\Sigma}_{\boldsymbol{\beta}_{n}} \right ) + \frac{1}{2} \text{tr}\left ( \frac{\partial^2 g_{nt}(\boldsymbol{\Gamma}_{n0})}{\partial \boldsymbol{\alpha}^2}  \boldsymbol{\Sigma}_{\alpha} \right ) \\ 
     \approx & \ln  \sum_{k \in C_{nt}} \exp(\boldsymbol{X}_{ntk,F} \boldsymbol{\mu}_{\alpha} +  \boldsymbol{X}_{ntk,R} \boldsymbol{\mu}_{\boldsymbol{\beta}_{n}})   \\
      & + \frac{1}{2}\text{tr}\left ( \left( \boldsymbol{X}_{nt,R}^{\top} \left( \text{diag}(\boldsymbol{p}_{nt0}) - \boldsymbol{p}_{nt0}\boldsymbol{p}_{nt0}^{\top}\right)\boldsymbol{X}_{nt,R} \right) \boldsymbol{\Sigma}_{\boldsymbol{\beta}_{n}} \right )\\
      & + \frac{1}{2}\text{tr}\left ( \left( \boldsymbol{X}_{nt,F}^{\top} \left( \text{diag}(\boldsymbol{p}_{nt0}) - \boldsymbol{p}_{nt0}\boldsymbol{p}_{nt0}^{\top}\right)\boldsymbol{X}_{nt,F} \right) \boldsymbol{\Sigma}_{\boldsymbol{\alpha}} \right),
\end{split} 
\end{equation}
where $p_{ntj,0} = \frac{\exp(\boldsymbol{X}_{ntj,F} \boldsymbol{\mu}_{\alpha} +  \boldsymbol{X}_{ntj,R} \boldsymbol{\mu}_{\boldsymbol{\beta}_{n}})}{\sum \limits_{k \in C_{nt}} \exp(\boldsymbol{X}_{ntk,R} \boldsymbol{\mu}_{\alpha} +  \boldsymbol{X}_{ntk,R} \boldsymbol{\mu}_{\boldsymbol{\beta}_{n}})}$ and $\boldsymbol{p}_{nt0} = \begin{bmatrix} p_{nt1,0} & \cdots & p_{ntJ,0} \end{bmatrix}$ is a row-vector of all $p_{ntj,0}$ in $C_{nt}$. 

\item Furthermore, QMC methods can be leveraged to approximate the E-LSE term by simulation:
\begin{equation}
\mathbb{E}_{q} \{g_{nt}(\boldsymbol{\Gamma}_{n})\} \approx
\frac{1}{D} \sum_{d = 1}^{D} 
 \ln  \sum_{k \in C_{nt}} \exp(\boldsymbol{X}_{ntk,F} \boldsymbol{\alpha}_{d} +  \boldsymbol{X}_{ntk,R} \boldsymbol{\beta}_{nd}),
\end{equation}
where
$\boldsymbol{\alpha}_{d} = \boldsymbol{\mu}_{\boldsymbol{\alpha}} + \text{chol}(\boldsymbol{\Sigma}_{\boldsymbol{\alpha}}) \boldsymbol{\xi}_{d,F}$
and 
$\boldsymbol{\beta}_{nd} = \boldsymbol{\mu}_{\boldsymbol{\beta}_{n}} + \text{chol}(\boldsymbol{\Sigma}_{\boldsymbol{\beta}_{n}}) \boldsymbol{\xi}_{nd,R}$. $\boldsymbol{\xi}_{d,F}$ and $\boldsymbol{\xi}_{nd,R}$ are points from a quasi-random sequence.

\item Finally, the modified Jensen's inequality can be used to define an alternative variational lower bound:
\begin{equation} \label{eq:MJI}
  \begin{split}
    \mathbb{E}_{q} \{g_{nt}(\boldsymbol{\Gamma}_{n})\} \leq & \sum_{k \in C_{nt}} a_{ntk}\boldsymbol{X}_{ntk}  \boldsymbol{\Gamma}_{n0} \\
    & + \ln\Bigg( \sum_{k \in C_{nt}} \exp \Bigg\{ \left(\boldsymbol{X}_{ntk} - \sum_{m \in C_{nt}}  a_{ntm}\boldsymbol{X}_{ntm}\right) \boldsymbol{\Gamma}_{n0} \\
    & + \frac{1}{2}\left(\boldsymbol{X}_{ntk} - \sum_{m \in C_{nt}}  a_{ntm}\boldsymbol{X}_{ntm}\right) \boldsymbol{V}_{\boldsymbol{\Gamma}_{n0}} \left(\boldsymbol{X}_{ntk} - \sum_{m \in C_{nt}}  a_{ntm}\boldsymbol{X}_{ntm}\right)^{\top} \Bigg\} \Bigg),
    \end{split}
\end{equation}
where 
\begin{equation} \label{eq:MJI_a}
a_{ntj} = \frac{\exp\left(\boldsymbol{X}_{ntj}  \boldsymbol{\Gamma}_{n0} + \frac{1}{2}\left(\boldsymbol{X}_{ntj} - 2\sum \limits_{m \in C_{nt}} a_{ntm}\boldsymbol{X}_{ntm}\right) \boldsymbol{V}_{\Gamma_{n0}}\boldsymbol{X}_{ntj}^{\top}\right)}{\sum \limits_{k \in C_{nt}} \exp\left(\boldsymbol{X}_{ntk}  \boldsymbol{\Gamma}_{n0} + \frac{1}{2}\left(\boldsymbol{X}_{ntk} - 2\sum \limits_{m \in C_{nt}}  a_{ntm}\boldsymbol{X}_{ntm}\right) \boldsymbol{V}_{\boldsymbol{\Gamma}_{n0}}\boldsymbol{X}_{ntk}^{\top} \right)} \quad \forall ntj
\end{equation} 
is an auxiliary variational parameter.
\end{enumerate}

Next, we outline the updating strategies for the nonconjugate variational factors:
\begin{enumerate}
\item With quasi-Newton (QN) methods \citep[e.g.][]{nocedal2006numerical}, updates for nonconjugate variational factors are obtained by maximizing the ELBO over the parameters of the variational factor in question. In that vein, updates for $q(\alpha)$ are given by
\begin{equation} \label{eq:alphaQN}
\begin{split}
\operatorname*{arg\,max}_{\boldsymbol{\mu}_{\boldsymbol{\alpha}}, \boldsymbol{\Sigma}_{\boldsymbol{\alpha}}} &
\Bigg \{
\sum_{n = 1}^{N} \sum_{t=1}^{T_{n}} \left ( \sum_{k \in C_{nt}} \left[ y_{ntk}(\boldsymbol{X}_{ntk,F} \boldsymbol{\mu}_{\alpha} +  \boldsymbol{X}_{ntk,R} \boldsymbol{\mu}_{\boldsymbol{\beta}_n} ) \right]
-\mathbb{E}_{q} \{g_{nt}(\boldsymbol{\Gamma}_{n})\} \right ) \\
& -\frac{1}{2} \text{tr} \left (\boldsymbol{\Xi}_{0}^{-1} \left (  \boldsymbol{\Sigma}_{\boldsymbol{\alpha}} + \boldsymbol{\mu}_{\boldsymbol{\alpha}}^{\top} \boldsymbol{\mu}_{\boldsymbol{\alpha}} \right)  \right ) + \boldsymbol{\mu}_{\boldsymbol{\alpha}}^{\top} \boldsymbol{\Xi}_{0}^{-1} \boldsymbol{\lambda}_{0}
  + \frac{1}{2}  \ln \vert \boldsymbol{\Sigma}_{\boldsymbol{\alpha}} \vert
\Bigg \},
\end{split}
\end{equation}
and updates for $q(\boldsymbol{\beta}_{n})$ are given by
\begin{equation}\label{eq:betaQN}
\begin{split}
\operatorname*{arg\,max}_{\boldsymbol{\mu}_{\boldsymbol{\beta}_{n}}, \boldsymbol{\Sigma}_{\boldsymbol{\beta}_{n}}} & 
\Bigg \{
\sum_{t=1}^{T_{n}} \left ( \sum_{k \in C_{nt}} \left[ y_{ntk}(\boldsymbol{X}_{ntk,F} \boldsymbol{\mu}_{\alpha} +  \boldsymbol{X}_{ntk,R} \boldsymbol{\mu}_{\boldsymbol{\beta}_n} ) \right]
-\mathbb{E}_{q} \{g_{nt}(\boldsymbol{\Gamma}_{n})\} \right ) \\
& -\frac{w}{2} \text{tr} \left (\boldsymbol{\Theta}^{-1} \boldsymbol{\Sigma}_{\boldsymbol{\beta}_{n}} \right )
- \frac{w}{2} \boldsymbol{\mu}_{\boldsymbol{\beta}_{n}}^{\top} \boldsymbol{\Theta}^{-1} \boldsymbol{\mu}_{\boldsymbol{\beta}_{n}} + w \boldsymbol{\mu}_{\boldsymbol{\beta}_{n}}^{\top} \boldsymbol{\Theta}^{-1} \boldsymbol{\mu}_{\boldsymbol{\zeta}}
  + \frac{1}{2}  \ln \vert \boldsymbol{\Sigma}_{\boldsymbol{\beta}_{n}} \vert
\Bigg \}.
\end{split}
\end{equation}
whereby the intractable E-LSE terms $\mathbb{E}_{q} \{g_{nt}(\boldsymbol{\Gamma}_{n})\}$ need to be replaced by an approximation or an alternative bound.

\item Nonconjugate variational message passing (NCVMP) admits the following fixed point updates for the parameters of $q(\boldsymbol{\alpha})$ and $q(\boldsymbol{\beta}_{n})$ \citep{wand2014fully}:
\begin{align}
\boldsymbol{\Sigma}_{\boldsymbol{\alpha}} & =  -\left [ 2\; \text{vec}^{-1} \left (
\nabla_{\text{vec}(\boldsymbol{\Sigma}_{\boldsymbol{\alpha}})} \left \{
\mathbb{E}_{q} \left \{ \ln P (\boldsymbol{y}_{1:N}, \boldsymbol{\theta}) \right\} \right \} \right )  \right ]^{-1} \label{eq:alphaNCVMP_Sigma} \\
\boldsymbol{\mu}_{\boldsymbol{\alpha}} & = \boldsymbol{\mu}_{\boldsymbol{\alpha}} + \boldsymbol{\Sigma}_{\boldsymbol{\alpha}} \left [
\nabla_{\boldsymbol{\mu}_{\boldsymbol{\alpha}}} \left \{ \mathbb{E}_{q} \left \{ \ln P (\boldsymbol{y}_{1:N}, \boldsymbol{\theta}) \right \} \right \} \right ], \label{eq:alphaNCVMP_mu} \\
\boldsymbol{\Sigma}_{\boldsymbol{\beta}_{n}} & =  -\left[2\; \text{vec}^{-1} \left ( 
\nabla_{\text{vec}(\boldsymbol{\Sigma}_{\boldsymbol{\beta}_{n}})} \left \{ \mathbb{E}_{q} \left \{ \ln P (\boldsymbol{y}_{1:N}, \boldsymbol{\theta}) \right \} \right\} \right ) \right]^{-1}, \label{eq:betaNCVMP_Sigma} \\
\boldsymbol{\mu}_{\boldsymbol{\beta}_{n}} & = \boldsymbol{\mu}_{\boldsymbol{\beta}_n} +  \boldsymbol{\Sigma}_{\boldsymbol{\beta}_n} \left[ 
\nabla_{\boldsymbol{\mu}_{\boldsymbol{\beta}_n}} \left \{ \mathbb{E}_{q}\left \{ \ln P (\boldsymbol{y}_{1:N}, \boldsymbol{\theta}) \right\} \right\}
\right]. \label{eq:betaNCVMP_mu} 
\end{align}
Here, if $\boldsymbol{B}$ is a matrix of dimension $K \times K$, then $b = \text{vec}(\boldsymbol{B})$ is a column-stacked vector of length $K^{2}$; $\text{vec}^{-1}(b) = \boldsymbol{B}$ reverses the operation. The term $\mathbb{E}_{q} \left \{ \ln P (\boldsymbol{y}_{1:N}, \boldsymbol{\theta}) \right \}$ is defined in expression \ref{eq:EofJoint} and involves intractable E-LSE terms, which need to be replaced by an approximation or bound. We derive the required gradient expressions (available upon request). We highlight that in contrast to QN methods, NCVMP does not guarantee that the ELBO increases after each iteration, because NCVMP involves only fixed point updates \citep{knowles2011non, wand2014fully}. However, NCVMP updates are substantially less costly than QN updates, as each NCVMP update involves only one function evaluation. 
\end{enumerate}

Algorithm \ref{algo_vb} succinctly summarizes the considered VB methods for posterior inference in MMNL models with a linear-in-parameters utility specification including both fixed and random utility parameters

\begin{algorithm}[h]
\textbf{Initialization:} \\
Set hyper-parameters: $\nu$, $A_{1:K}$, $\boldsymbol{\mu}_{0}$, $\boldsymbol{\Sigma}_{0}$, $\boldsymbol{\lambda}_{0}$, $\boldsymbol{\Xi}_{0}$; \\
Provide starting values: $\boldsymbol{\mu}_{\boldsymbol{\zeta}}$, $\boldsymbol{\Sigma}_{\boldsymbol{\zeta}}$, $\boldsymbol{\mu}_{\boldsymbol{\beta}_{1:N}}$, $\boldsymbol{\Sigma}_{\boldsymbol{\beta}_{1:N}}$, $d_{1:K}$; \\
\If{VB-QN-MJI or VB-NCVMP-MJI}{
	Set $a_{ntj} = \frac{1}{\vert C_{nt} \vert}$ $\forall ntj$;\\
	}
\textbf{Coordinate ascent:} \\
\If{VB-QN-QMC}{
	Generate standard normal quasi-random sequences: $\boldsymbol{\xi}_{1:D}$, $\boldsymbol{\delta}_{1:N,1:D}$; \\
	}
$c = \frac{\nu + K}{2}$; $w = \nu + N + K - 1$; $\boldsymbol{\Theta} = 2 \nu \text{diag} \left ( \frac{c}{\boldsymbol{d}} \right ) + N \boldsymbol{\Sigma_{\zeta}} + \sum_{n = 1}^{N} \left ( \boldsymbol{\Sigma}_{\boldsymbol{\beta}_{n}} + (\boldsymbol{\mu}_{\boldsymbol{\beta}_{n}} - \boldsymbol{\mu}_{\boldsymbol{\zeta}}) (\boldsymbol{\mu}_{\boldsymbol{\beta}_{n}} - \boldsymbol{\mu}_{\boldsymbol{\zeta}})^{\top} \right)$;\\
\While{not converged}{
	\If{VB-QN-$\Delta$ or VB-QN-QMC or VB-QN-MJI}{
		Update $\boldsymbol{\mu}_{\boldsymbol{\alpha}}$, $\boldsymbol{\Sigma}_{\boldsymbol{\alpha}}$ using equation \ref{eq:alphaQN}; \\
		Update $\boldsymbol{\mu}_{\boldsymbol{\beta}_{n}}$, $\boldsymbol{\Sigma}_{\boldsymbol{\beta}_{n}}$ for $\forall n$ using equation \ref{eq:betaQN}; \\
	}
	\If{VB-NCVMP-$\Delta$ or VB-NCVMP-MJI}{
	   	Update $\boldsymbol{\mu}_{\boldsymbol{\alpha}}$, $\boldsymbol{\Sigma}_{\boldsymbol{\alpha}}$ using equations \ref{eq:alphaNCVMP_mu} and \ref{eq:alphaNCVMP_Sigma}; \\	   
	   	Update $\boldsymbol{\mu}_{\boldsymbol{\beta}_{n}}$, $\boldsymbol{\Sigma}_{\boldsymbol{\beta}_{n}}$ for $\forall n$ using equations \ref{eq:betaNCVMP_mu} and \ref{eq:betaNCVMP_Sigma}; \\
	}
	$\boldsymbol{\Sigma}_{\boldsymbol{\zeta}} = \left ( \boldsymbol{\Sigma}_{0}^{-1} + N w \boldsymbol{\Theta}^{-1} \right )^{-1}$;\\
	$\boldsymbol{\mu}_{\boldsymbol{\zeta}} = \boldsymbol{\Sigma}_{\boldsymbol{\zeta}} \left ( \boldsymbol{\Sigma}_{0}^{-1} \boldsymbol{\mu}_{0} + w \boldsymbol{\Theta}^{-1} \sum_{n = 1}^{N} \boldsymbol{\mu}_{\boldsymbol{\beta}_{n}}  \right )$; \\
	$\boldsymbol{\Theta} = 2 \nu \text{diag} \left ( \frac{c}{\boldsymbol{d}} \right ) + N \boldsymbol{\Sigma_{\zeta}} + \sum_{n = 1}^{N} \left ( \boldsymbol{\Sigma}_{\boldsymbol{\beta}_{n}} + (\boldsymbol{\mu}_{\boldsymbol{\beta}_{n}} - \boldsymbol{\mu}_{\boldsymbol{\zeta}}) (\boldsymbol{\mu}_{\boldsymbol{\beta}_{n}} - \boldsymbol{\mu}_{\boldsymbol{\zeta}})^{\top} \right)$; \\
	$d_{k} = \frac{1}{A_{k}^{2}} + \nu w \left ( \boldsymbol{\Theta}^{-1} \right )_{kk}$ $\forall k $; \\
	\If{VB-QN-MJI or VB-NCVMP-MJI}{
		Update $a_{ntj}$ for $\forall ntj$ using equation \ref{eq:MJI_a}; 
	   }
}
\caption{Pseudo-code representations of variational Bayes methods for posterior inference in MMNL models with a linear-in-parameters utility specification including both fixed and random utility parameters} \label{algo_vb}
\end{algorithm}

\section{Simulation evaluation} \label{sec:simeval}

\subsection{Data and experimental setup}

For the simulation study, we devise a semi-synthetic data generating process (DGP), under which the choice sets and population parameters are based on real data from a stated choice study on consumer preferences for alternative fuel vehicles in Germany \citep{achtnicht2012german}. The real data comprise 3,588 observations from 598 individuals. In the original study, respondents were presented with six choice sets, each of which consisted of seven alternatives, which in turn were characterized by six attributes, namely fuel type and propulsion technology (gasoline, diesel, hybrid, LPG/CNG, biofuel, hydrogen, electric), purchase price, operating costs, engine power, CO\textsubscript{2} emissions and fuel availability.

We generate the semi-synthetic choice data as follows: Decision-makers are assumed to be utility maximizers and to evaluate alternatives based on the utility specification $U_{ntj} = \boldsymbol{X}_{ntj,F} \boldsymbol{\alpha} + \boldsymbol{X}_{ntj,R} \boldsymbol{\beta}_{n} + \epsilon_{ntj}$. Here, $n \in \{ 1, \ldots, N \}$ indexes decision-makers, $t \in \{1, \ldots, T \}$ indexes choice occasions, and $j \in \{1, \ldots, 7 \}$ indexes alternatives. $\boldsymbol{X}_{ntj,F}$ is a row-vector of attributes for which tastes $\boldsymbol{\alpha}$ are invariant across decision-makers (gasoline, hybrid, LPG/CNG, biofuel, hydrogen, electric, purchase price); $\boldsymbol{X}_{ntj,R}$ is a row-vector of attributes for which tastes $\boldsymbol{\beta}_{n}$ are individual-specific (operating costs, engine power, CO\textsubscript{2} emissions, fuel availability). The choice sets $\boldsymbol{X}_{nt,1:7}$ with $\boldsymbol{X}_{ntj} = \begin{bmatrix} \boldsymbol{X}_{ntj,F} & \boldsymbol{X}_{ntj,R} \end{bmatrix}$ are drawn from the real data with equal probability and with replacement. $\epsilon_{ntj}$ is a stochastic disturbance sampled from $\text{Gumbel}(0,1)$. The individual-specific taste parameters are drawn from a multivariate normal distribution, i.e. $\boldsymbol{\beta}_{n} \sim \text{N}(\boldsymbol{\zeta}, \boldsymbol{\Omega})$ for $n = 1, \ldots, N$ with $\boldsymbol{\Omega} = \text{diag}(\boldsymbol{\sigma}) \boldsymbol{\Psi} \text{diag}(\boldsymbol{\sigma})$, where $\boldsymbol{\sigma}$ is a standard deviation vector, and $\boldsymbol{\Psi}$ is a correlation matrix. The values of $\boldsymbol{\alpha}$, $\boldsymbol{\zeta}$, and $\boldsymbol{\sigma}$ are based on maximum simulated likelihood point estimates of the parameters of a mixed multinomial logit model fit to the real data. The scale of the population-level parameters is set such that the error rate is approximately 50\%, i.e. in 50\% of the cases decision-makers deviate from the deterministically-best alternative due to the stochastic utility component. 

We consider four experimental scenarios: In scenarios 1 and 2, the fixed taste parameters and their corresponding attributes are omitted from the utility specification in the DGP, and only the individual-specific parameters are estimated. In scenarios 3 and 4, the full utility specification is used in the DGP, and both sets of taste parameters are estimated. Furthermore, the degree of correlation among individual-specific taste parameters is relatively low in scenarios 1 and 3, whereas it is relatively high in scenarios 2 and 4. In Appendix \ref{A:true}, we enumerate the values of $\boldsymbol{\alpha}$, $\boldsymbol{\zeta}$, $\boldsymbol{\sigma}$, and $\boldsymbol{\Psi}$ for each experimental scenario. In each scenario, $N$ takes a value in $\{500, 2000\}$, and $T$ takes a value in $\{5, 10\}$. For each experimental scenario and combination of $N$ and $T$, we consider 20 replications, whereby the data for each replication are generated based on a different random seed.   

\subsection{Accuracy assessment}

We employ two performance metrics to assess the accuracy of the estimation approaches:
\begin{enumerate}
\item To evaluate how the estimation approaches perform at recovering parameters, we calculate the root mean square error (RMSE) for selected parameters, namely for the invariant parameter vector $\boldsymbol{\alpha}$, the mean vector $\boldsymbol{\zeta}$, the unique elements of the covariance matrix $\boldsymbol{\Omega}_{U}$ and the matrix of individual-specific taste parameters $\boldsymbol{\beta}_{1:N}$. Given collections of parameters $\boldsymbol{\theta}$ and their estimates $\hat{\boldsymbol{\theta}}$, RMSE is defined as
\begin{equation}
\text{RMSE}(\boldsymbol{\theta}) = \sqrt{ \frac{1}{M} (\hat{\boldsymbol{\theta}} - \boldsymbol{\theta})^{\top} (\hat{\boldsymbol{\theta}} - \boldsymbol{\theta})},
\end{equation}
where $M$ denotes the total number of scalar parameters collected in $\boldsymbol{\theta}$. For MSLE, point estimates of $\boldsymbol{\alpha}$, $\boldsymbol{\zeta}$ and $\boldsymbol{\Omega}_{U}$ are directly obtained. Point estimates of $\boldsymbol{\beta}_{1:N}$ are given by the following conditional expectation \citep{revelt1999customer}:
\begin{equation} \label{eq:indSpecBeta}
\hat{\boldsymbol{\beta}}_{n} =
\mathbb{E} \{ \boldsymbol{\beta}_{n} \vert \boldsymbol{y}_{n}, \boldsymbol{X}_{n}, \hat{\boldsymbol{\alpha}}, \hat{\boldsymbol{\zeta}}, \hat{\boldsymbol{\Omega}} \} =
\frac{\int \boldsymbol{\beta}_{n} P(\boldsymbol{y}_{n} \vert \boldsymbol{X}_{n}, \hat{\boldsymbol{\alpha}},  \boldsymbol{\beta}_{n}) f(\boldsymbol{\beta}_{n} \vert \hat{\boldsymbol{\zeta}}, \hat{\boldsymbol{\Omega}}) d \boldsymbol{\beta}_{n}}{\int P(\boldsymbol{y}_{n} \vert \boldsymbol{X}_{n}, \hat{\boldsymbol{\alpha}},  \boldsymbol{\beta}_{n}) f(\boldsymbol{\beta}_{n} \vert \hat{\boldsymbol{\zeta}}, \hat{\boldsymbol{\Omega}}) d \boldsymbol{\beta}_{n}}, 
\quad n = 1, \ldots, N.
\end{equation}
The integrals in expression \ref{eq:indSpecBeta} are intractable and are thus simulated using 10,000 pseudo-random draws. For MCMC, estimates of the parameters of interest are given by the means of the respective posterior draws. For VB, we have $\hat{\boldsymbol{\alpha}} = \boldsymbol{\mu}_{\boldsymbol{\alpha}}$, $\hat{\boldsymbol{\zeta}} = \boldsymbol{\mu}_{\boldsymbol{\zeta}}$, $\hat{\boldsymbol{\Omega}} = \frac{1}{w - K - 1} \boldsymbol{\Theta}$ and $\hat{\boldsymbol{\beta}}_{n} = \boldsymbol{\mu}_{\boldsymbol{\beta}_{n}}$ for $ n = 1, \ldots, N$. As we are interested in evaluating how well the estimation methods perform at recovering the distributions of the realized tastes, we use the sample mean $\boldsymbol{\zeta}_{0} = \frac{1}{N} \sum_{n = 1}^{N} \boldsymbol{\beta}_{n}$ and the sample covariance $\boldsymbol{\Omega}_{0} = \frac{1}{N} \sum_{n = 1}^{N} (\boldsymbol{\beta}_{n} - \boldsymbol{\zeta}_{0}) (\boldsymbol{\beta}_{n} - \boldsymbol{\zeta}_{0})^{\top}$ of the draws of the individual-specific parameters $\boldsymbol{\beta}_{1:N}$ as true values for $\boldsymbol{\zeta}$ and $\boldsymbol{\Omega}$, respectively. 

\item To evaluate the out-of-sample predictive accuracy of the estimation approaches, we compute the total variation distance \citep[TVD;][]{braun2010variational} between the true and the estimated predictive choice distributions for a validation sample, which we generate along with each training sample. Each validation sample is based on the same DGP as its respective training sample, whereby the number of decision-makers is set to 25 and the number of observations per decision-maker is set to one. The true predictive choice distribution for a choice set $C_{nt}$ with attributes $\boldsymbol{X}_{nt}^{*}$ from the validation sample is given by
\begin{equation}
P_{\text{true}}(y_{nt}^{*} \vert \boldsymbol{X}_{nt}^{*}) = 
\int P(y_{nt}^{*} = j \vert \boldsymbol{X}_{nt}^{*}, \boldsymbol{\alpha}, \boldsymbol{\beta}) f(\boldsymbol{\beta} \vert \boldsymbol{\zeta}, \boldsymbol{\Omega}) d \boldsymbol{\beta}.
\end{equation}
This integration is not tractable and is therefore simulated using 1,000,000 pseudo-random draws from the true heterogeneity distribution $\text{N}(\boldsymbol{\zeta}, \boldsymbol{\Omega})$. The corresponding estimated predictive choice distribution is 
\begin{equation}
\hat{P} (y_{nt}^{*} \vert \boldsymbol{X}_{nt}^{*}, \boldsymbol{y}) = 
\int \int \int
\left ( \int P(y_{nt}^{*} \vert \boldsymbol{X}_{nt}^{*}, \boldsymbol{\alpha}, \boldsymbol{\beta}) f(\boldsymbol{\beta} \vert \boldsymbol{\zeta}, \boldsymbol{\Omega}) d \boldsymbol{\beta} \right ) p(\boldsymbol{\alpha}, \boldsymbol{\zeta}, \boldsymbol{\Omega} \vert \boldsymbol{y}) d \boldsymbol{\alpha} d \boldsymbol{\zeta} d \boldsymbol{\Omega}.
\end{equation}
The estimated posterior predictive distribution can be computed via Monte Carlo integration. For MCMC, $p(\boldsymbol{\alpha}, \boldsymbol{\zeta}, \boldsymbol{\Omega} \vert \boldsymbol{y})$ is given by the empirical distribution of the posterior draws. For VB, $p(\boldsymbol{\alpha}, \boldsymbol{\zeta}, \boldsymbol{\Omega} \vert \boldsymbol{y})$ is replaced by the estimated variational distribution $q(\boldsymbol{\alpha}) q(\boldsymbol{\zeta}) q(\boldsymbol{\Omega})$. We note that the posterior predictive choice distribution is a quintessentially Bayesian quantity, which accounts for the uncertainty in the parameter estimates by marginalizing the predictive distribution over the posterior distribution of the parameters. By contrast, frequentist predictions are based on point estimates. In the current application, we mimic the posterior predictive distribution for MSLE by marginalizing the predictive distribution over the asymptotic distribution $\text{N}(\hat{\boldsymbol{\varphi}}, \text{var} \{ \hat{\boldsymbol{\varphi}} \})$ of the parameter estimates. Here $\hat{\boldsymbol{\varphi}}$ denotes the point estimate of $\{ \boldsymbol{\alpha}, \boldsymbol{\zeta}, \text{chol}(\boldsymbol{\Omega}) \}$, and $\text{var} \{ \hat{\boldsymbol{\varphi}} \}$ denotes the corresponding asymptotic variance-covariance of $\hat{\boldsymbol{\varphi}}$. $\text{var} \{ \hat{\boldsymbol{\varphi}} \}$ is the Cram\'er-Rao bound, which we approximate by evaluating the inverse of the negative Hessian matrix of the log-likelihood function at the point estimates.\footnote{To be precise, we consider the Hessian approximation returned by the Broyden-Fletcher-Goldfarb-Shanno algorithm \citep{nocedal2006numerical}.} For VB and MSLE, we take 500 pseudo-random draws for $\{ \boldsymbol{\alpha}, \boldsymbol{\zeta}, \boldsymbol{\Omega} \}$ from $q(\boldsymbol{\alpha}) q(\boldsymbol{\zeta}) q(\boldsymbol{\Omega})$ and $\text{N}(\hat{\boldsymbol{\varphi}}, \text{var} \{ \hat{\boldsymbol{\varphi}} \})$; for MCMC, we use 20,000 draws from $p(\boldsymbol{\alpha}, \boldsymbol{\zeta}, \boldsymbol{\Omega} \vert \boldsymbol{y})$. For MCMC, a larger number of draws is necessary, as the posterior draws are not independent. For all methods, we use 10,000 i.i.d draws for $\boldsymbol{\beta}$. TVD is then given by 
\begin{equation}
\text{TVD} = \frac{1}{2} \sum_{j \in C_{nt}} \left | 
P_{\text{true}}(y_{nt}^{*} = j \vert \boldsymbol{X}_{nt}^{*})  - 
\hat{P} (y_{nt}^{*} = j \vert \boldsymbol{X}_{nt}^{*}, \boldsymbol{y})  \right |.
\end{equation}
For succinctness, we calculate averages across decision-makers and choice sets.
\end{enumerate}

\subsection{Implementation details}

We implement all estimation approaches described above by writing our own Python code\footnote{The Python code is publicly available at \url{https://github.com/RicoKrueger/bayes_mxl}.} and make an effort that the implementations of the different estimators are as similar as possible to allow for fair comparisons of estimation times. The computation of the simulated log-likelihood for MSLE and all sampling steps of the MCMC algorithm can be fully vectorized. However, VB estimation necessarily involves loops to update the variational factors pertaining to the individual-specific taste parameters. For MSLE, choice probabilities are simulated using 1,000 simulation draws generated via the Modified Latin Hypercube Sampling method \citep{hess2006use}. For VB-QN-QMC, we use 64 simulation draws generated via the same method; we also explored larger numbers of simulation draws (128, 256) for VB-QN-QMC but found that increases in the number of simulation draws resulted in prohibitive estimation times. For MSLE and VB-QN, we employ the Broyden-Fletcher-Goldfarb-Shanno algorithm \citep{nocedal2006numerical} included in Python's SciPy library \citep{jones2001open} to carry out the numerical optimizations; the default settings of the algorithm are used and analytical or simulated gradients are supplied. To assure positive-definiteness of the covariance matrices, all numerical optimizations are in fact performed with respect to the Cholesky factors of the covariance matrices. For MCMC, the sampler is executed with two parallel Markov chains and 100,000 iterations for each chain, whereby the initial 50,000 iterations of each chain are discarded for burn-in. After burn-in, every fifth draw is retained to reduce the amount of autocorrelation in the chains. For the VB methods, we apply the same stopping criterion as \citet{tan2017stochastic}: We define $\boldsymbol{\vartheta} = \begin{bmatrix} \boldsymbol{\alpha}^{\top} & \boldsymbol{\zeta}^{\top} & \text{diag}(\boldsymbol{\Psi})^{\top} & \boldsymbol{d}^{\top} \end{bmatrix}^{\top}$ and let $\vartheta_{i}^{(\tau)}$ denote the $i$th element of $\boldsymbol{\vartheta}$ at iteration $\tau$. We terminate the iterative coordinate ascent algorithm, when $\delta^{(\tau)} = \operatorname*{arg\,max}_{i} \frac{\vert \vartheta_{i}^{(\tau + 1)} - \vartheta_{i}^{(\tau)}\vert }{\vert \vartheta_{i}^{(\tau)} \vert} < 0.005$. As $\delta^{(\tau)}$ can fluctuate, $\boldsymbol{\vartheta}^{(\tau)}$ is substituted by its average over the last five iterations. The simulation experiments are conducted on the Katana high performance computing cluster at the Faculty of Science, UNSW Australia.

\subsection{Results}

Tables \ref{table_results_S1} to \ref{table_results_S4} enumerate the results for scenarios 1 to 4, respectively. Each table gives the means and the standard errors of the considered performance metrics for 20 replications under different combinations of sample sizes $N \in \{500,2000\}$ and choice occasions per decision-maker $T \in \{5,10\}$. In principle, a statistical testing procedure such as ANOVA could be used to compare the performance metrics of the different methods. Here, we will simply compare mean estimates, as the standard errors are generally small. 

First, we examine the impact of the sample size $N$ and the number of choice occasions per decision-maker $T$ on the performance of the estimation methods. For all methods, the mean RMSE of $\boldsymbol{\alpha}$, $\boldsymbol{\zeta}$, $\boldsymbol{\Omega}_{U}$ and $\boldsymbol{\beta}_{1:N}$ as well as the mean TVD decrease with the sample size $N$ and the number of occasions $T$. These findings numerically validate the consistency of VB methods \citep[see][]{wang2018frequentist}. In our subsequent discussion, we only make explicit mention of numerical results for $\{N=2000,T=10\}$, as the comparative performance of the estimation methods is generally consistent across all combinations of $N$ and $T$. 

All methods recover the mean vector $\boldsymbol{\zeta}$ and the individual-specific parameters $\boldsymbol{\beta}_{1:N}$ equally well in the considered scenarios. For example, the mean RMSE values of $\boldsymbol{\zeta}$ fall into tight intervals of $[0.0181, 0.0281]$, $[0.0191, 0.0252]$, $[0.0246, 0.0286]$, and $[0.0246,0.0291]$. Likewise, the corresponding ranges for $\boldsymbol{\beta}_{1:N}$ are $[0.7198, 0.7257]$, $[0.6929, 0.6959]$, $[0.7217, 0.7254]$,  and $[0.6955, 0.6982]$. Furthermore, the results of scenarios 3 and 4 show that the fixed parameters $\boldsymbol{\alpha}$ are also recovered equally well by the considered methods. Narrow ranges of the corresponding mean RMSE values across all methods in both scenarios support this observation: $[0.0269, 0.0277]$, $[0.0298, 0.0307]$. 

With the exception of the VB methods relying on the MJI-based alternative variational lower bound, all methods perform equally well at recovering the covariance matrix $\boldsymbol{\Omega}$. Excluding VB-QN-MJI and VB-NCVMP-MJI, the mean RMSE values of $\boldsymbol{\Omega}_{U}$ lie in narrow ranges of $[0.0568, 0.0800]$, $[0.0570, 0.0665]$, $[0.0711, 0.0736]$ and $[0.0572, 0.0692]$, whereas the mean RMSE values of $\boldsymbol{\Omega}_{U}$ for VB-QN-MJI and VB-NCVMP-MJI are substantially larger. Upon close inspection of the simulation results, it can be seen that that the magnitudes of the relative differences in the mean RMSE value of $\boldsymbol{\Omega}_{U}$ between the MJI-based VB methods and the other methods increase, as $N$ rises. For all methods, the recovery of $\boldsymbol{\Omega}_{U}$ ameliorates, as the number of choice occasions per decision-maker increases. Furthermore, we observe that the degree of correlation does not affect the quality of the estimation of all methods. 

Next, we compare the predictive accuracy of the estimation methods. With the exception of VB-QN-MJI and VB-NCVMP-MJI, the estimation approaches perform equally well at prediction. The lower predictive accuracy of MJI-based methods can be attributed to a less accurate recovery of the covariance matrix $\boldsymbol{\Omega}$. In the majority of the considered experimental conditions, the MJI-based VB methods perform noticeably worse than the competing methods, which implies that the alternative variational lower bound defined with the help of the modified Jensen's inequality affords less accurate inferences than the analytical and simulation-based E-LSE approximations. This finding is consistent with \citet{depraetere2017comparison}. We also observe that the TVD proxy for MSLE is comparable to the actual TVD calculated for the Bayesian methods.

Finally, we contrast the computational efficiency of the estimation methods. For VB, we observe that NCVMP updates are substantially faster than QN updates at virtually no compromises in parameter recovery and predictive accuracy. In contrast to earlier studies \citep{braun2010variational, depraetere2017comparison}, we do not find that the QN-based VB methods are faster than MCMC, even though we use similar numbers of draws for the posterior simulations. A possible explanation for this discrepancy is that earlier studies rely on the bayesm \citep{rossi2012bayesian} package for R to carry out the MCMC estimations, whereas we develop our own efficient Python implementation. Of the considered VB methods, VB-NCVMP-$\Delta$ performs best at balancing fast estimation times, acceptable parameter recovery and good predictive accuracy. Across the considered experimental conditions, VB-NCVMP-$\Delta$ is on average between 1.7 to 16.2 times faster than MCMC and MSLE, while performing nearly as well at prediction and parameter recovery. Whereas earlier studies reported occasional convergence issues for the delta-method-based E-LSE approximation \citep{depraetere2017comparison, tan2017stochastic}, we encountered no such issues in the current simulation study.

\begin{landscape}
\begin{table}[h]
\scriptsize
\centering
\ra{1.2}
\begin{tabular}{@{} l 
S[table-format=4.1] S[table-format=4.1]  c 
S[table-format=1.4] S[table-format=1.4]  c 
S[table-format=1.4] S[table-format=1.4]  c 
S[table-format=1.4] S[table-format=1.4]  c  
S[table-format=1.4] S[table-format=1.4]  @{}} 
\toprule

& 
\multicolumn{2}{c}{\textbf{Estimation time}} & &
\multicolumn{2}{c}{\textbf{$\mbox{RMSE}(\boldsymbol{\zeta})$}} & &
\multicolumn{2}{c}{\textbf{$\mbox{RMSE}(\boldsymbol{\Omega}_{U})$}} & &
\multicolumn{2}{c}{\textbf{$\mbox{RMSE}(\boldsymbol{\beta}_{1:N})$}} & &
\multicolumn{2}{c}{\textbf{TVD [\%]}} \\
\cmidrule{2-3} \cmidrule{5-6} \cmidrule{8-9} \cmidrule{11-12}  \cmidrule{14-15} 

& 
\textbf{Mean} & \textbf{Std. err.} & &
\textbf{Mean} & \textbf{Std. err.} & &
\textbf{Mean} & \textbf{Std. err.} & &
\textbf{Mean} & \textbf{Std. err.} & &
\textbf{Mean} & \textbf{Std. err.} \\
\midrule

$N = 500$; $T = 5$ \\
\quad MSLE &   213.2 &   5.8 & &0.0723 & 0.0055 & &0.2297 & 0.0201 & &0.8485 & 0.0039 & &0.3607 & 0.0171 \\
\quad MCMC &   203.3 &   4.8 & &0.0713 & 0.0059 & &0.2189 & 0.0192 & &0.8454 & 0.0040 & &0.3479 & 0.0171 \\
\quad VB-QN-$\Delta$ &   591.5 &  29.9 & &0.0722 & 0.0061 & &0.2131 & 0.0202 & &0.8438 & 0.0042 & &0.3505 & 0.0174 \\
\quad VB-QN-QMC &  4593.4 & 227.7 & &0.0695 & 0.0055 & &0.1947 & 0.0193 & &0.8420 & 0.0041 & &0.3459 & 0.0171 \\
\quad VB-QN-MJI &   424.4 &  23.6 & &0.0817 & 0.0079 & &0.2546 & 0.0098 & &0.8478 & 0.0036 & &0.3680 & 0.0164 \\
\quad VB-NCVMP-$\Delta$ &    45.3 &   2.5 & &0.0720 & 0.0061 & &0.2150 & 0.0197 & &0.8442 & 0.0041 & &0.3510 & 0.0174 \\
\quad VB-NCVMP-MJI &    26.6 &   1.6 & &0.0828 & 0.0080 & &0.2598 & 0.0096 & &0.8484 & 0.0036 & &0.3691 & 0.0162 \\
$N = 500$; $T = 10$ \\
\quad MSLE &   507.8 &  21.5 & &0.0468 & 0.0049 & &0.1388 & 0.0187 & &0.7315 & 0.0052 & &0.2620 & 0.0170 \\
\quad MCMC &   284.6 &   5.9 & &0.0448 & 0.0039 & &0.1240 & 0.0113 & &0.7248 & 0.0035 & &0.2553 & 0.0155 \\
\quad VB-QN-$\Delta$ &   678.8 &  25.3 & &0.0463 & 0.0040 & &0.1218 & 0.0097 & &0.7241 & 0.0035 & &0.2566 & 0.0153 \\
\quad VB-QN-QMC &  3194.6 & 139.7 & &0.0444 & 0.0040 & &0.1149 & 0.0098 & &0.7234 & 0.0035 & &0.2545 & 0.0154 \\
\quad VB-QN-MJI &   391.9 &  21.9 & &0.0456 & 0.0044 & &0.1552 & 0.0123 & &0.7274 & 0.0037 & &0.2509 & 0.0162 \\
\quad VB-NCVMP-$\Delta$ &    44.5 &   2.1 & &0.0463 & 0.0040 & &0.1244 & 0.0101 & &0.7245 & 0.0035 & &0.2567 & 0.0153 \\
\quad VB-NCVMP-MJI &    22.8 &   1.4 & &0.0457 & 0.0045 & &0.1587 & 0.0123 & &0.7278 & 0.0037 & &0.2512 & 0.0161 \\
$N = 2000$; $T = 5$ \\
\quad MSLE &   815.5 &  22.1 & &0.0285 & 0.0027 & &0.1041 & 0.0079 & &0.8330 & 0.0014 & &0.1578 & 0.0090 \\
\quad MCMC &   459.1 &   7.2 & &0.0280 & 0.0027 & &0.1052 & 0.0086 & &0.8329 & 0.0015 & &0.1569 & 0.0090 \\
\quad VB-QN-$\Delta$ &  2168.5 & 129.6 & &0.0327 & 0.0037 & &0.1136 & 0.0081 & &0.8334 & 0.0015 & &0.1693 & 0.0092 \\
\quad VB-QN-QMC & 16418.2 & 854.3 & &0.0288 & 0.0027 & &0.1056 & 0.0069 & &0.8328 & 0.0014 & &0.1606 & 0.0095 \\
\quad VB-QN-MJI &  1538.6 &  79.7 & &0.0530 & 0.0031 & &0.2573 & 0.0065 & &0.8442 & 0.0015 & &0.2163 & 0.0099 \\
\quad VB-NCVMP-$\Delta$ &   175.6 &   8.0 & &0.0321 & 0.0036 & &0.1203 & 0.0081 & &0.8338 & 0.0015 & &0.1700 & 0.0090 \\
\quad VB-NCVMP-MJI &    93.0 &   5.2 & &0.0557 & 0.0033 & &0.2643 & 0.0063 & &0.8450 & 0.0015 & &0.2178 & 0.0100 \\
$N = 2000$; $T = 10$ \\
\quad MSLE &  2185.4 & 112.8 & &0.0200 & 0.0017 & &0.0800 & 0.0144 & &0.7257 & 0.0039 & &0.1462 & 0.0110 \\
\quad MCMC &   739.5 &  18.1 & &0.0181 & 0.0016 & &0.0575 & 0.0036 & &0.7198 & 0.0017 & &0.1345 & 0.0105 \\
\quad VB-QN-$\Delta$ &  2675.8 &  98.3 & &0.0191 & 0.0016 & &0.0575 & 0.0032 & &0.7198 & 0.0017 & &0.1365 & 0.0109 \\
\quad VB-QN-QMC & 13011.3 & 442.9 & &0.0186 & 0.0017 & &0.0568 & 0.0035 & &0.7198 & 0.0017 & &0.1340 & 0.0105 \\
\quad VB-QN-MJI &  1572.7 &  37.2 & &0.0279 & 0.0023 & &0.1290 & 0.0050 & &0.7233 & 0.0017 & &0.1487 & 0.0110 \\
\quad VB-NCVMP-$\Delta$ &   196.6 &   7.0 & &0.0190 & 0.0016 & &0.0585 & 0.0032 & &0.7198 & 0.0017 & &0.1367 & 0.0109 \\
\quad VB-NCVMP-MJI &    96.7 &   2.8 & &0.0281 & 0.0023 & &0.1328 & 0.0051 & &0.7236 & 0.0017 & &0.1494 & 0.0110 \\

\midrule
\multicolumn{15}{l}{
\begin{minipage}[t]{1.0 \textwidth}
Note: 
$\boldsymbol{\zeta}$, and $\boldsymbol{\Omega}_{U}$: mean vector and unique elements of covariance matrix; 
$\boldsymbol{\beta}_{1:N}$: matrix of individual-specific taste parameters;
TVD: total variation distance between true and predicted choice probabilities for a validation sample.
\end{minipage}} \\
\bottomrule
\end{tabular}
\caption{Results for scenario 1 (low correlation, only random taste parameters) } \label{table_results_S1}
\end{table}
\end{landscape}

\begin{landscape}
\begin{table}[h]
\scriptsize
\centering
\ra{1.2}
\begin{tabular}{@{} l 
S[table-format=4.1] S[table-format=4.1]  c 
S[table-format=1.4] S[table-format=1.4]  c 
S[table-format=1.4] S[table-format=1.4]  c 
S[table-format=1.4] S[table-format=1.4]  c  
S[table-format=1.4] S[table-format=1.4]  @{}} 
\toprule

& 
\multicolumn{2}{c}{\textbf{Estimation time}} & &
\multicolumn{2}{c}{\textbf{$\mbox{RMSE}(\boldsymbol{\zeta})$}} & &
\multicolumn{2}{c}{\textbf{$\mbox{RMSE}(\boldsymbol{\Omega}_{U})$}} & &
\multicolumn{2}{c}{\textbf{$\mbox{RMSE}(\boldsymbol{\beta}_{1:N})$}} & &
\multicolumn{2}{c}{\textbf{TVD [\%]}} \\
\cmidrule{2-3} \cmidrule{5-6} \cmidrule{8-9} \cmidrule{11-12}  \cmidrule{14-15} 

& 
\textbf{Mean} & \textbf{Std. err.} & &
\textbf{Mean} & \textbf{Std. err.} & &
\textbf{Mean} & \textbf{Std. err.} & &
\textbf{Mean} & \textbf{Std. err.} & &
\textbf{Mean} & \textbf{Std. err.} \\
\midrule

$N = 500$; $T = 5$ \\
\quad MSLE &   202.0 &   6.7 & &0.0612 & 0.0076 & &0.1857 & 0.0145 & &0.8155 & 0.0034 & &0.3322 & 0.0192 \\
\quad MCMC &   198.6 &   3.8 & &0.0617 & 0.0074 & &0.1743 & 0.0152 & &0.8131 & 0.0032 & &0.3205 & 0.0198 \\
\quad VB-QN-$\Delta$ &   567.0 &  23.2 & &0.0661 & 0.0080 & &0.1630 & 0.0125 & &0.8119 & 0.0034 & &0.3283 & 0.0214 \\
\quad VB-QN-QMC &  4168.7 & 172.7 & &0.0603 & 0.0070 & &0.1555 & 0.0097 & &0.8108 & 0.0031 & &0.3196 & 0.0190 \\
\quad VB-QN-MJI &   487.4 &  30.3 & &0.0682 & 0.0059 & &0.2408 & 0.0118 & &0.8149 & 0.0030 & &0.3348 & 0.0201 \\
\quad VB-NCVMP-$\Delta$ &    37.5 &   2.0 & &0.0657 & 0.0081 & &0.1622 & 0.0123 & &0.8119 & 0.0034 & &0.3286 & 0.0214 \\
\quad VB-NCVMP-MJI &    28.4 &   2.1 & &0.0688 & 0.0059 & &0.2458 & 0.0113 & &0.8153 & 0.0030 & &0.3350 & 0.0200 \\
$N = 500$; $T = 10$ \\
\quad MSLE &   540.9 &  29.3 & &0.0457 & 0.0039 & &0.1471 & 0.0142 & &0.7072 & 0.0033 & &0.2673 & 0.0199 \\
\quad MCMC &   291.5 &   9.6 & &0.0461 & 0.0036 & &0.1326 & 0.0088 & &0.7021 & 0.0028 & &0.2608 & 0.0212 \\
\quad VB-QN-$\Delta$ &   677.0 &  43.0 & &0.0483 & 0.0036 & &0.1256 & 0.0082 & &0.7015 & 0.0028 & &0.2638 & 0.0213 \\
\quad VB-QN-QMC &  3097.9 & 123.2 & &0.0461 & 0.0036 & &0.1271 & 0.0087 & &0.7016 & 0.0028 & &0.2613 & 0.0219 \\
\quad VB-QN-MJI &   440.5 &  26.2 & &0.0454 & 0.0039 & &0.1420 & 0.0101 & &0.7019 & 0.0026 & &0.2599 & 0.0206 \\
\quad VB-NCVMP-$\Delta$ &    39.6 &   2.6 & &0.0478 & 0.0036 & &0.1251 & 0.0083 & &0.7015 & 0.0028 & &0.2636 & 0.0213 \\
\quad VB-NCVMP-MJI &    21.4 &   1.1 & &0.0453 & 0.0040 & &0.1456 & 0.0100 & &0.7022 & 0.0026 & &0.2599 & 0.0206 \\
$N = 2000$; $T = 5$ \\
\quad MSLE &   880.8 &  19.0 & &0.0392 & 0.0046 & &0.1053 & 0.0077 & &0.8076 & 0.0015 & &0.1764 & 0.0114 \\
\quad MCMC &   463.3 &  11.1 & &0.0395 & 0.0046 & &0.1049 & 0.0072 & &0.8072 & 0.0015 & &0.1763 & 0.0113 \\
\quad VB-QN-$\Delta$ &  2068.2 &  77.1 & &0.0441 & 0.0047 & &0.1004 & 0.0072 & &0.8067 & 0.0015 & &0.1858 & 0.0098 \\
\quad VB-QN-QMC & 14491.5 & 482.3 & &0.0403 & 0.0043 & &0.0984 & 0.0073 & &0.8065 & 0.0015 & &0.1803 & 0.0116 \\
\quad VB-QN-MJI &  2149.2 & 126.4 & &0.0559 & 0.0036 & &0.2401 & 0.0083 & &0.8147 & 0.0017 & &0.2174 & 0.0104 \\
\quad VB-NCVMP-$\Delta$ &   132.8 &   4.5 & &0.0429 & 0.0046 & &0.1012 & 0.0075 & &0.8067 & 0.0015 & &0.1847 & 0.0099 \\
\quad VB-NCVMP-MJI &   139.9 &   8.5 & &0.0566 & 0.0035 & &0.2433 & 0.0078 & &0.8150 & 0.0017 & &0.2178 & 0.0103 \\
$N = 2000$; $T = 10$ \\
\quad MSLE &  2244.3 & 105.6 & &0.0193 & 0.0019 & &0.0665 & 0.0079 & &0.6953 & 0.0025 & &0.1417 & 0.0086 \\
\quad MCMC &   739.2 &  13.2 & &0.0194 & 0.0016 & &0.0577 & 0.0034 & &0.6930 & 0.0019 & &0.1305 & 0.0084 \\
\quad VB-QN-$\Delta$ &  2123.1 & 130.4 & &0.0208 & 0.0016 & &0.0585 & 0.0036 & &0.6929 & 0.0019 & &0.1312 & 0.0087 \\
\quad VB-QN-QMC & 11027.8 & 246.9 & &0.0191 & 0.0019 & &0.0570 & 0.0035 & &0.6930 & 0.0019 & &0.1351 & 0.0082 \\
\quad VB-QN-MJI &  1746.2 &  60.3 & &0.0235 & 0.0024 & &0.1178 & 0.0048 & &0.6953 & 0.0020 & &0.1375 & 0.0077 \\
\quad VB-NCVMP-$\Delta$ &   153.3 &   6.3 & &0.0204 & 0.0016 & &0.0619 & 0.0034 & &0.6931 & 0.0019 & &0.1310 & 0.0087 \\
\quad VB-NCVMP-MJI &    73.0 &   2.7 & &0.0252 & 0.0025 & &0.1270 & 0.0050 & &0.6959 & 0.0020 & &0.1391 & 0.0076 \\

\midrule
\multicolumn{15}{l}{
\begin{minipage}[t]{1.0 \textwidth}
Note: For an explanation of the column headers see Table \ref{table_results_S1}.
\end{minipage}} \\
\bottomrule
\end{tabular}
\caption{Results for scenario 2 (high correlation, only random taste parameters)} \label{table_results_S2}
\end{table}
\end{landscape}

\begin{landscape}
\begin{table}[h]
\scriptsize
\centering
\ra{1.2}
\begin{tabular}{@{} l 
S[table-format=4.1] S[table-format=4.1]  c 
S[table-format=1.4] S[table-format=1.4]  c 
S[table-format=1.4] S[table-format=1.4]  c 
S[table-format=1.4] S[table-format=1.4]  c 
S[table-format=1.4] S[table-format=1.4]  c  
S[table-format=1.4] S[table-format=1.4]  @{}} 
\toprule

& 
\multicolumn{2}{c}{\textbf{Estimation time}} & &
\multicolumn{2}{c}{\textbf{$\mbox{RMSE}(\boldsymbol{\alpha})$}} & &
\multicolumn{2}{c}{\textbf{$\mbox{RMSE}(\boldsymbol{\zeta})$}} & &
\multicolumn{2}{c}{\textbf{$\mbox{RMSE}(\boldsymbol{\Omega}_{U})$}} & &
\multicolumn{2}{c}{\textbf{$\mbox{RMSE}(\boldsymbol{\beta}_{1:N})$}} & &
\multicolumn{2}{c}{\textbf{TVD [\%]}} \\
\cmidrule{2-3} \cmidrule{5-6} \cmidrule{8-9} \cmidrule{11-12}  \cmidrule{14-15}  \cmidrule{17-18} 

& 
\textbf{Mean} & \textbf{Std. err.} & &
\textbf{Mean} & \textbf{Std. err.} & &
\textbf{Mean} & \textbf{Std. err.} & &
\textbf{Mean} & \textbf{Std. err.} & &
\textbf{Mean} & \textbf{Std. err.} & &
\textbf{Mean} & \textbf{Std. err.} \\
\midrule

$N = 500$; $T = 5$ \\
\quad MSLE &   279.9 &    6.4 & &0.0752 & 0.0054 & &0.0587 & 0.0039 & &0.1927 & 0.0150 & &0.8450 & 0.0035 & &0.4259 & 0.0196 \\
\quad MCMC &   319.3 &    6.3 & &0.0750 & 0.0055 & &0.0594 & 0.0037 & &0.1918 & 0.0174 & &0.8436 & 0.0039 & &0.4204 & 0.0198 \\
\quad VB-QN-$\Delta$ &  4997.5 &  291.5 & &0.0782 & 0.0059 & &0.0685 & 0.0049 & &0.1830 & 0.0159 & &0.8431 & 0.0035 & &0.4220 & 0.0199 \\
\quad VB-QN-QMC &  5052.5 &  284.7 & &0.0754 & 0.0056 & &0.0596 & 0.0039 & &0.1668 & 0.0157 & &0.8410 & 0.0037 & &0.4169 & 0.0207 \\
\quad VB-QN-MJI &  1943.4 &  104.4 & &0.0732 & 0.0054 & &0.0611 & 0.0048 & &0.2492 & 0.0099 & &0.8466 & 0.0035 & &0.4347 & 0.0188 \\
\quad VB-NCVMP-$\Delta$ &   141.9 &    4.0 & &0.0779 & 0.0059 & &0.0678 & 0.0048 & &0.1842 & 0.0155 & &0.8432 & 0.0035 & &0.4221 & 0.0196 \\
\quad VB-NCVMP-MJI &    45.3 &    2.9 & &0.0732 & 0.0054 & &0.0620 & 0.0049 & &0.2548 & 0.0097 & &0.8473 & 0.0035 & &0.4351 & 0.0184 \\
$N = 500$; $T = 10$ \\
\quad MSLE &   712.4 &   28.8 & &0.0595 & 0.0035 & &0.0477 & 0.0038 & &0.1332 & 0.0132 & &0.7352 & 0.0053 & &0.3316 & 0.0140 \\
\quad MCMC &   469.9 &    8.8 & &0.0594 & 0.0034 & &0.0443 & 0.0037 & &0.1144 & 0.0061 & &0.7295 & 0.0043 & &0.3283 & 0.0131 \\
\quad VB-QN-$\Delta$ &  6050.3 &  361.7 & &0.0599 & 0.0034 & &0.0455 & 0.0037 & &0.1138 & 0.0059 & &0.7293 & 0.0043 & &0.3334 & 0.0136 \\
\quad VB-QN-QMC &  3917.0 &  158.8 & &0.0599 & 0.0034 & &0.0444 & 0.0037 & &0.1094 & 0.0057 & &0.7288 & 0.0043 & &0.3326 & 0.0126 \\
\quad VB-QN-MJI &  1985.1 &   78.1 & &0.0597 & 0.0034 & &0.0486 & 0.0038 & &0.1605 & 0.0059 & &0.7328 & 0.0042 & &0.3367 & 0.0130 \\
\quad VB-NCVMP-$\Delta$ &   161.3 &    3.5 & &0.0599 & 0.0034 & &0.0455 & 0.0037 & &0.1155 & 0.0061 & &0.7295 & 0.0043 & &0.3318 & 0.0137 \\
\quad VB-NCVMP-MJI &    44.8 &    1.6 & &0.0598 & 0.0035 & &0.0487 & 0.0038 & &0.1639 & 0.0059 & &0.7332 & 0.0042 & &0.3366 & 0.0135 \\
$N = 2000$; $T = 5$ \\
\quad MSLE &  1166.4 &   33.2 & &0.0461 & 0.0031 & &0.0373 & 0.0036 & &0.1086 & 0.0081 & &0.8402 & 0.0017 & &0.2236 & 0.0107 \\
\quad MCMC &   818.8 &   18.9 & &0.0466 & 0.0032 & &0.0370 & 0.0035 & &0.1089 & 0.0085 & &0.8400 & 0.0017 & &0.2222 & 0.0110 \\
\quad VB-QN-$\Delta$ & 22339.5 & 1470.5 & &0.0466 & 0.0034 & &0.0409 & 0.0033 & &0.1152 & 0.0074 & &0.8403 & 0.0018 & &0.2288 & 0.0101 \\
\quad VB-QN-QMC & 18354.7 &  784.1 & &0.0457 & 0.0031 & &0.0358 & 0.0037 & &0.1067 & 0.0067 & &0.8397 & 0.0016 & &0.2226 & 0.0112 \\
\quad VB-QN-MJI &  8786.7 &  485.6 & &0.0484 & 0.0040 & &0.0548 & 0.0050 & &0.2643 & 0.0068 & &0.8514 & 0.0015 & &0.2551 & 0.0116 \\
\quad VB-NCVMP-$\Delta$ &   499.8 &   21.8 & &0.0467 & 0.0033 & &0.0405 & 0.0033 & &0.1217 & 0.0075 & &0.8407 & 0.0018 & &0.2291 & 0.0102 \\
\quad VB-NCVMP-MJI &   174.4 &   10.4 & &0.0484 & 0.0040 & &0.0575 & 0.0051 & &0.2711 & 0.0064 & &0.8523 & 0.0015 & &0.2572 & 0.0118 \\
$N = 2000$; $T = 10$ \\
\quad MSLE &  3064.9 &  164.9 & &0.0269 & 0.0023 & &0.0248 & 0.0027 & &0.0716 & 0.0045 & &0.7225 & 0.0015 & &0.1648 & 0.0072 \\
\quad MCMC &  1497.3 &   25.9 & &0.0274 & 0.0027 & &0.0253 & 0.0026 & &0.0726 & 0.0045 & &0.7218 & 0.0015 & &0.1651 & 0.0077 \\
\quad VB-QN-$\Delta$ & 28820.1 & 1457.0 & &0.0273 & 0.0026 & &0.0267 & 0.0025 & &0.0719 & 0.0045 & &0.7219 & 0.0016 & &0.1661 & 0.0075 \\
\quad VB-QN-QMC & 16896.0 &  510.0 & &0.0272 & 0.0025 & &0.0246 & 0.0027 & &0.0711 & 0.0039 & &0.7217 & 0.0015 & &0.1651 & 0.0074 \\
\quad VB-QN-MJI &  7960.8 &  360.2 & &0.0276 & 0.0020 & &0.0285 & 0.0036 & &0.1276 & 0.0077 & &0.7250 & 0.0018 & &0.1733 & 0.0076 \\
\quad VB-NCVMP-$\Delta$ &   574.7 &   30.8 & &0.0273 & 0.0026 & &0.0266 & 0.0025 & &0.0736 & 0.0049 & &0.7220 & 0.0016 & &0.1669 & 0.0077 \\
\quad VB-NCVMP-MJI &   164.9 &    7.0 & &0.0277 & 0.0020 & &0.0286 & 0.0035 & &0.1319 & 0.0079 & &0.7254 & 0.0019 & &0.1743 & 0.0078 \\

\midrule
\multicolumn{18}{l}{
\begin{minipage}[t]{1.0 \textwidth}
Note: 
$\boldsymbol{\alpha}$: fixed parameter vector; 
$\boldsymbol{\zeta}$ and $\boldsymbol{\Omega}_{U}$: mean vector and unique elements of covariance matrix; $\boldsymbol{\beta}_{1:N}$: matrix of individual-specific taste parameters;
TVD: total variation distance between true and predicted choice probabilities for a validation sample.
\end{minipage}} \\
\bottomrule
\end{tabular}
\caption{Results for scenario 3 (low correlation, combination of fixed and random taste parameters)} \label{table_results_S3}
\end{table}
\end{landscape}

\begin{landscape}
\begin{table}[h]
\scriptsize
\centering
\ra{1.2}
\begin{tabular}{@{} l 
S[table-format=4.1] S[table-format=4.1]  c 
S[table-format=1.4] S[table-format=1.4]  c 
S[table-format=1.4] S[table-format=1.4]  c 
S[table-format=1.4] S[table-format=1.4]  c 
S[table-format=1.4] S[table-format=1.4]  c  
S[table-format=1.4] S[table-format=1.4]  @{}} 
\toprule

& 
\multicolumn{2}{c}{\textbf{Estimation time}} & &
\multicolumn{2}{c}{\textbf{$\mbox{RMSE}(\boldsymbol{\alpha})$}} & &
\multicolumn{2}{c}{\textbf{$\mbox{RMSE}(\boldsymbol{\zeta})$}} & &
\multicolumn{2}{c}{\textbf{$\mbox{RMSE}(\boldsymbol{\Omega}_{U})$}} & &
\multicolumn{2}{c}{\textbf{$\mbox{RMSE}(\boldsymbol{\beta}_{1:N})$}} & &
\multicolumn{2}{c}{\textbf{TVD [\%]}} \\
\cmidrule{2-3} \cmidrule{5-6} \cmidrule{8-9} \cmidrule{11-12}  \cmidrule{14-15}  \cmidrule{17-18} 

& 
\textbf{Mean} & \textbf{Std. err.} & &
\textbf{Mean} & \textbf{Std. err.} & &
\textbf{Mean} & \textbf{Std. err.} & &
\textbf{Mean} & \textbf{Std. err.} & &
\textbf{Mean} & \textbf{Std. err.} & &
\textbf{Mean} & \textbf{Std. err.} \\
\midrule

$N = 500$; $T = 5$ \\
\quad MSLE &   294.6 &    8.3 & &0.0873 & 0.0077 & &0.0708 & 0.0072 & &0.2115 & 0.0220 & &0.8125 & 0.0044 & &0.4548 & 0.0225 \\
\quad MCMC &   321.6 &    4.7 & &0.0874 & 0.0079 & &0.0721 & 0.0075 & &0.2034 & 0.0237 & &0.8081 & 0.0044 & &0.4430 & 0.0233 \\
\quad VB-QN-$\Delta$ &  5291.6 &  339.9 & &0.0881 & 0.0081 & &0.0814 & 0.0079 & &0.2219 & 0.0286 & &0.8098 & 0.0053 & &0.4468 & 0.0246 \\
\quad VB-QN-QMC &  4747.0 &  267.8 & &0.0867 & 0.0079 & &0.0708 & 0.0074 & &0.1879 & 0.0232 & &0.8067 & 0.0045 & &0.4386 & 0.0234 \\
\quad VB-QN-MJI &  2364.4 &  165.2 & &0.0892 & 0.0075 & &0.0700 & 0.0066 & &0.2017 & 0.0104 & &0.8060 & 0.0032 & &0.4462 & 0.0225 \\
\quad VB-NCVMP-$\Delta$ &   120.1 &    6.3 & &0.0880 & 0.0081 & &0.0806 & 0.0079 & &0.2172 & 0.0280 & &0.8095 & 0.0052 & &0.4465 & 0.0244 \\
\quad VB-NCVMP-MJI &    51.4 &    3.1 & &0.0892 & 0.0075 & &0.0706 & 0.0066 & &0.2050 & 0.0105 & &0.8063 & 0.0032 & &0.4480 & 0.0228 \\
$N = 500$; $T = 10$ \\
\quad MSLE &   783.8 &   26.7 & &0.0527 & 0.0029 & &0.0556 & 0.0044 & &0.1586 & 0.0151 & &0.7098 & 0.0040 & &0.3573 & 0.0151 \\
\quad MCMC &   476.4 &   10.2 & &0.0531 & 0.0028 & &0.0500 & 0.0046 & &0.1322 & 0.0089 & &0.6995 & 0.0026 & &0.3463 & 0.0173 \\
\quad VB-QN-$\Delta$ &  6562.5 &  296.5 & &0.0531 & 0.0029 & &0.0493 & 0.0049 & &0.1244 & 0.0088 & &0.6983 & 0.0026 & &0.3466 & 0.0174 \\
\quad VB-QN-QMC &  3970.1 &  119.1 & &0.0531 & 0.0029 & &0.0493 & 0.0047 & &0.1274 & 0.0087 & &0.6986 & 0.0027 & &0.3456 & 0.0168 \\
\quad VB-QN-MJI &  2288.0 &  176.3 & &0.0532 & 0.0030 & &0.0525 & 0.0045 & &0.1479 & 0.0104 & &0.7004 & 0.0027 & &0.3477 & 0.0178 \\
\quad VB-NCVMP-$\Delta$ &   135.8 &    5.8 & &0.0531 & 0.0029 & &0.0497 & 0.0048 & &0.1273 & 0.0087 & &0.6987 & 0.0026 & &0.3461 & 0.0175 \\
\quad VB-NCVMP-MJI &    43.8 &    2.3 & &0.0532 & 0.0030 & &0.0529 & 0.0046 & &0.1518 & 0.0106 & &0.7008 & 0.0027 & &0.3481 & 0.0179 \\
$N = 2000$; $T = 5$ \\
\quad MSLE &  1260.5 &   26.0 & &0.0404 & 0.0030 & &0.0406 & 0.0044 & &0.1142 & 0.0063 & &0.8095 & 0.0013 & &0.2238 & 0.0082 \\
\quad MCMC &   789.8 &   20.2 & &0.0404 & 0.0031 & &0.0414 & 0.0045 & &0.1126 & 0.0062 & &0.8090 & 0.0013 & &0.2257 & 0.0086 \\
\quad VB-QN-$\Delta$ & 19041.7 & 1369.5 & &0.0395 & 0.0032 & &0.0477 & 0.0052 & &0.1087 & 0.0089 & &0.8091 & 0.0013 & &0.2288 & 0.0095 \\
\quad VB-QN-QMC & 16885.7 &  662.6 & &0.0397 & 0.0031 & &0.0402 & 0.0042 & &0.0981 & 0.0047 & &0.8082 & 0.0013 & &0.2260 & 0.0084 \\
\quad VB-QN-MJI & 12503.1 &  796.1 & &0.0470 & 0.0038 & &0.0487 & 0.0043 & &0.2190 & 0.0105 & &0.8143 & 0.0017 & &0.2570 & 0.0087 \\
\quad VB-NCVMP-$\Delta$ &   383.9 &   20.3 & &0.0396 & 0.0031 & &0.0467 & 0.0050 & &0.1093 & 0.0080 & &0.8090 & 0.0013 & &0.2299 & 0.0098 \\
\quad VB-NCVMP-MJI &   237.4 &   15.8 & &0.0470 & 0.0038 & &0.0495 & 0.0043 & &0.2229 & 0.0101 & &0.8146 & 0.0017 & &0.2588 & 0.0087 \\
$N = 2000$; $T = 10$ \\
\quad MSLE &  2885.7 &  144.7 & &0.0302 & 0.0021 & &0.0256 & 0.0026 & &0.0692 & 0.0082 & &0.6982 & 0.0025 & &0.1834 & 0.0069 \\
\quad MCMC &  1439.6 &   34.0 & &0.0307 & 0.0025 & &0.0250 & 0.0022 & &0.0601 & 0.0036 & &0.6957 & 0.0015 & &0.1804 & 0.0060 \\
\quad VB-QN-$\Delta$ & 24572.4 & 1430.7 & &0.0299 & 0.0022 & &0.0260 & 0.0023 & &0.0581 & 0.0034 & &0.6955 & 0.0014 & &0.1805 & 0.0060 \\
\quad VB-QN-QMC & 15307.7 &  349.9 & &0.0298 & 0.0022 & &0.0246 & 0.0022 & &0.0572 & 0.0033 & &0.6955 & 0.0014 & &0.1779 & 0.0065 \\
\quad VB-QN-MJI &  9008.9 &  500.4 & &0.0304 & 0.0020 & &0.0282 & 0.0025 & &0.1071 & 0.0048 & &0.6975 & 0.0014 & &0.1836 & 0.0064 \\
\quad VB-NCVMP-$\Delta$ &   493.3 &   23.2 & &0.0299 & 0.0021 & &0.0257 & 0.0023 & &0.0603 & 0.0033 & &0.6957 & 0.0014 & &0.1810 & 0.0060 \\
\quad VB-NCVMP-MJI &   142.9 &    7.4 & &0.0303 & 0.0020 & &0.0291 & 0.0027 & &0.1161 & 0.0050 & &0.6981 & 0.0014 & &0.1843 & 0.0064 \\

\midrule
\multicolumn{18}{l}{
\begin{minipage}[t]{1.0 \textwidth}
Note: For an explanation of the column headers see Table \ref{table_results_S3}.
\end{minipage}} \\
\bottomrule
\end{tabular}
\caption{Results for scenario 4 (high correlation, combination of fixed and random taste parameters)} \label{table_results_S4}
\end{table}
\end{landscape}

\section{Conclusions} \label{sec:concl}

This study extends several variational Bayes (VB) methods to allow for posterior inference in mixed multinomial logit (MMNL) models with a linear-in-parameters utility specification involving both taste parameters that vary normally across decision-makers as well as taste parameters that are invariant across decision-makers. In addition, extensive simulation-based evaluations provide new evidence into the finite-sample properties and the predictive accuracy of VB methods for MMNL in comparison with Markov chain Monte Carlo (MCMC) methods and maximum simulated likelihood estimation (MSLE). Our findings suggest that VB with nonconjugate variational message passing and a delta-method-based approximation of the expectation of log-sum of exponential (E-LSE) term (VB-NCVMP-$\Delta$) is an attractive alternative to MCMC and MSLE for fast and scalable estimation of MMNL models. The substantial gains in computational efficiency come at practically no compromises in parameter recovery and predictive accuracy. 

There are several directions in which future research can build on the work presented in the current paper. First, VB methods for posterior inference in MMNL models are currently limited to MMNL models with normal mixing distributions and utility specifications in preference space. Extending VB methods to accommodate more flexible parametric, nonparametric, and semiparametric mixing distributions as well as utility specifications in willingness-to-pay space is an immediate step to support the use of VB methods in empirical applications. Second, VB methods can be devised for extended discrete choice models \citep{walker2001extended} such as the integrated choice and latent variable model. As excessive estimation times continue to represent a bottleneck in empirical applications of such advanced discrete choice models, VB methods could facilitate the use of these and other behaviourally-rich models in novel contexts and applications. Third, we have shown that VB methods perform reasonably well at recovering individual-level parameters and lend themselves well to applications in which fast predictions are paramount. Thus, our analysis may inform the development of online estimation procedures that could enable near real time learning and prediction of individual preferences. Fourth, to further accelerate VB estimation for large datasets, stochastic variational inference methods can be leveraged \citep[see][]{hoffman2013stochastic, tan2017stochastic}.

Adaptations of VB to other discrete choice models, new contexts and applications may benefit from fundamental advancements in the underlying VB procedure. First, in this paper, we have considered extensions to a standard VB approach, which relies on the KL divergence and the mean-field assumption. While computationally-convenient, the KL divergence is known to be a relatively loose bound, which may in turn lead to an underestimation of posterior variances \citep[see][and the literature cited therein]{zhang2018advances}. Thus, other probability divergences such as $\alpha$- and $f$-divergences \citep[also see][for an overview]{zhang2018advances} may be explored in future work. The mean-field assumption is  computationally convenient, but it restricts the flexibility of the variational distribution to an extent that the exact posterior can never be assumed by its variational approximation \citep{zhang2018advances}. The quality of the variational distribution may be improved by injecting structure into the formulation of the variational distribution. This may be achieved by explicitly recognizing that some parameters are hierarchically dependent \citep[e.g.][]{ranganath2016hierarchical}. Second, Markov chain variational inference \citep[MCVI;][]{salimans2015markov, wolf2016variational} seeks to combine the conceptual benefits of MCMC and VB, i.e. i.e. accurate inferences and fast estimation, respectively. Developing an MCVI method for MMNL is another potential direction for future research. Third, enhancements in the analytical and simulation-based approximation of E-LSE could lead to further improvements in the computational efficiency and quality of the VB methods. Improvements in computational efficiency may also be realized by leveraging advancements in technical computing soft- and hardware.

Finally, another avenue for future research is to contrast the VB methods considered in the current study with other emerging analytical approximation methods proposed in the frequentist context such as the Maximum Approximate Composite Marginal Likelihood (MACML) approach \citep{bhat2014new, bhat2018new, bhat2011simulation, patil2017simulation}.

\section*{Acknowledgements}
We are grateful to Chandra Bhat, Abdul Pinjari and two anonymous reviewers for their critical assessment of our work.
We would also like to thank Martin Achtnicht for sharing the stated choice data, Tim Hillel for help with the Python implementation, and Naveen Sunder for many helpful comments on the first draft. 
PB and RAD are thankful to the National Science Foundation CAREER Award CBET-1253475 for financially supporting this research. PB is also thankful to Prof. Joan Walker and Prof. Kenneth Train for their guidance during his visit to UC Berkeley under the Exchange Scholar Program. 
RK and THR acknowledge financial support from the Australian Research Council (DE170101346).
This research includes computations using the Linux computational cluster Katana supported by the Faculty of Science, UNSW Australia.

\section*{Author contribution statement}
PB: conception and design, model extension and implementation (VB-NCVMP), data preparation and analysis, manuscript writing and editing. \\
RK: conception and design, model extension and implementation (VB-QN), model implementation (MCMC, MSLE, VB-NCVMP), data preparation and analysis, manuscript writing and editing. \\
MB: conception and design, manuscript editing, supervision. \\
RAD: conception and design, manuscript editing, supervision. \\
THR: conception and design, manuscript editing, supervision. \\

\newpage
\bibliographystyle{apalike}
\bibliography{bibliography.bib}

\newpage
\begin{appendices}
	
\section{Optimal densities of conjugate variational factors} \label{A:optQ}

\subsection{\texorpdfstring{$q^{*}(a_{k})$}{q*(a_k)}}

\begin{equation}
\begin{split}
q^{*}(a_{k}) 
& \propto \exp \mathbb{E}_{- a_{k}} \left \{ \ln P(a_{k} \vert s,  r_{k}) + \ln P(\boldsymbol{\Omega} \vert \omega, \boldsymbol{B}) \right \} \\
& \propto \exp \mathbb{E}_{- a_{k}} \left \{ (s - 1) \ln a_{k} - r_{k} a_{k} + \frac{\omega}{2} \ln \boldsymbol{B}_{kk} - \frac{1}{2} \boldsymbol{B}_{kk} \left ( \boldsymbol{\Omega}^{-1} \right )_{kk} \right \} \\
& \propto \exp \left \{ \left ( \frac{\nu + K}{2} - 1 \right ) \ln a_{k} - \left ( r_{k} + \nu \mathbb{E}_{- a_{k}} \left \{ \left ( \boldsymbol{\Omega}^{-1} \right )_{kk} \right \} \right ) a_{k} \right \} \\
& \propto \text{Gamma}(c, d_{k}),
\end{split}
\end{equation}
where 
$c = \frac{\nu + K}{2}$ and 
$d_{k} = \frac{1}{A_{k}^{2}} + \nu \mathbb{E}_{- a_{k}} \left \{ \left ( \boldsymbol{\Omega}^{-1} \right )_{kk} \right \}$. 
Furthermore, we note that $\mathbb{E} a_{k} = \frac{c}{d_{k}}$ and $\boldsymbol{d} = \begin{pmatrix} d_{1} & \dots & d_{K} \end{pmatrix}^{\top}$.

\subsection{\texorpdfstring{$q^{*}(\boldsymbol{\zeta})$}{q*(zeta)}}

\begin{equation}
\begin{split}
q^{*}(\boldsymbol{\zeta}) 
& \propto \exp \mathbb{E}_{- \boldsymbol{\zeta}} \left \{ \ln P(\boldsymbol{\zeta} \vert \boldsymbol{\mu}_{0},\boldsymbol{\Sigma}_{0}) + \sum_{n=1}^{N} \ln P(\boldsymbol{\beta}_{n} \vert \boldsymbol{\zeta}, \boldsymbol{\Omega}) \right \} \\
& \propto \exp \mathbb{E}_{- \boldsymbol{\zeta}} \left \{ 
- \frac{1}{2} \boldsymbol{\zeta}^{\top} \boldsymbol{\Sigma}_{0}^{-1} \boldsymbol{\zeta} + \boldsymbol{\zeta}^{\top} \boldsymbol{\Sigma}_{0}^{-1} \boldsymbol{\mu}_{0}
- \frac{N}{2} \boldsymbol{\zeta}^{\top} \boldsymbol{\Omega}^{-1} \boldsymbol{\zeta} + \sum_{n = 1}^{N} \boldsymbol{\zeta}^{\top} \boldsymbol{\Omega}^{-1} \boldsymbol{\beta}_{n}
\right \} \\
& \propto \exp \left \{ 
- \frac{1}{2} \left (
\boldsymbol{\zeta}^{\top} \left ( \boldsymbol{\Sigma}_{0}^{-1} + N \mathbb{E}_{- \boldsymbol{\zeta}} \left \{ \boldsymbol{\Omega}^{-1} \right \} \right ) \boldsymbol{\zeta}
- 2 \boldsymbol{\zeta}^{\top} \left ( \boldsymbol{\Sigma}_{0}^{-1} \boldsymbol{\mu}_{0} + \mathbb{E}_{- \boldsymbol{\zeta}} \left \{  \boldsymbol{\Omega}^{-1} \right \} \sum_{n = 1}^{N} \mathbb{E}_{- \boldsymbol{\zeta}} \boldsymbol{\beta}_{n} \right )
\right )
\right \} \\
& \propto \text{Normal}(\boldsymbol{\mu}_{\boldsymbol_{\zeta}}, \boldsymbol{\Sigma}_{\boldsymbol{\zeta}}),
\end{split}
\end{equation}
where 
$\boldsymbol{\Sigma}_{\boldsymbol{\zeta}} = \left ( \boldsymbol{\Sigma}_{0}^{-1} + N \mathbb{E}_{- \boldsymbol{\zeta}} \left \{ \boldsymbol{\Omega}^{-1} \right \} \right )^{-1}$ and 
$\boldsymbol{\mu}_{\boldsymbol_{\zeta}} = \boldsymbol{\Sigma}_{\boldsymbol{\zeta}} \left ( \boldsymbol{\Sigma}_{0}^{-1} \boldsymbol{\mu}_{0} + \mathbb{E}_{- \boldsymbol{\zeta}} \left \{  \boldsymbol{\Omega}^{-1} \right \} \sum_{n = 1}^{N} \mathbb{E}_{- \boldsymbol{\zeta}}  \boldsymbol{\beta}_{n}  \right )$. 
Furthermore, we note that 
$\mathbb{E} \boldsymbol{\zeta} = \boldsymbol{\mu}_{\boldsymbol_{\zeta}}$ and $\mathbb{E} \boldsymbol{\beta}_{n} = \boldsymbol{\mu}_{\boldsymbol{\beta}_{n}}$.

\subsection{\texorpdfstring{$q^{*}(\boldsymbol{\Omega})$}{q*(Omega)}}

\begin{equation} 
\begin{split}
q^{*}(\boldsymbol{\Omega}) 
& \propto \exp \mathbb{E}_{- \boldsymbol{\Omega}} \left \{ \ln P(\boldsymbol{\Omega} \vert \omega, \boldsymbol{B}) + \sum_{n=1}^{N} \ln P(\boldsymbol{\beta}_{n} \vert \boldsymbol{\zeta}, \boldsymbol{\Omega}) \right \} \\
& \propto \exp \mathbb{E}_{- \boldsymbol{\Omega}} \left \{ 
- \frac{\omega + K + 1}{2} \ln \vert \boldsymbol{\Omega} \vert - \frac{1}{2} \text{tr} \left ( \boldsymbol{B} \boldsymbol{\Omega}^{-1} \right )
- \frac{N}{2} \ln \vert \boldsymbol{\Omega} \vert - \frac{1}{2} \sum_{n = 1}^{N} (\boldsymbol{\beta}_{n} - \boldsymbol{\zeta})^{\top} \boldsymbol{\Omega}^{-1} (\boldsymbol{\beta}_{n} - \boldsymbol{\zeta})
\right \} \\
& = \exp \left \{ 
- \frac{\omega + N + K + 1}{2} \ln \vert \boldsymbol{\Omega} \vert - \frac{1}{2} \text{tr} \left ( \boldsymbol{\Omega}^{-1} \mathbb{E}_{- \boldsymbol{\Omega}} \left \{ \boldsymbol{B} + 
 \sum_{n = 1}^{N} (\boldsymbol{\beta}_{n} - \boldsymbol{\zeta}) (\boldsymbol{\beta}_{n} - \boldsymbol{\zeta})^{\top} \right \} \right )
\right \} \\
& \propto \text{IW}(w, \boldsymbol{\Theta}),
\end{split}
\end{equation}
where 
$w = \nu + N + K - 1$ and 
$\boldsymbol{\Theta} = 
2 \nu \text{diag} \left ( \frac{c}{\boldsymbol{d}} \right )
+ N \boldsymbol{\Sigma_{\zeta}}
+ \sum_{n = 1}^{N} \left (
\boldsymbol{\Sigma}_{\boldsymbol{\beta}_{n}} + (\boldsymbol{\mu}_{\boldsymbol{\beta}_{n}} - \boldsymbol{\mu}_{\boldsymbol{\zeta}}) (\boldsymbol{\mu}_{\boldsymbol{\beta}_{n}} - \boldsymbol{\mu}_{\boldsymbol{\zeta}})^{\top}
 \right)
$. We use $\mathbb{E}\left(\boldsymbol{\beta}_{n} \boldsymbol{\beta}_{n}^{\top}\right) = \boldsymbol{\mu}_{\boldsymbol{\beta}_{n}} \boldsymbol{\mu}_{\boldsymbol{\beta}_{n}}^{\top} + \boldsymbol{\Sigma}_{\boldsymbol{\beta}_{n}}$ and $\mathbb{E}\left(\boldsymbol{\zeta}\boldsymbol{\zeta}^{\top}\right) = \boldsymbol{\mu}_{\boldsymbol{\zeta}} \boldsymbol{\mu}_{\boldsymbol{\zeta}}^{\top} + \boldsymbol{\Sigma}_{\boldsymbol{\zeta}}$. 
Furthermore, we note that 
$\mathbb{E} \{ \boldsymbol{\Omega}^{-1} \} = w \boldsymbol{\Theta}^{-1}$ 
and
$\mathbb{E} \{ \ln \vert \boldsymbol{\Omega} \vert \} = \ln \vert  \boldsymbol{\Theta} \vert + C$, where $C$ is a constant.

\section{True population parameters for the simulation study} \label{A:true}

\begin{align}
\boldsymbol{\alpha} & =
\begin{bmatrix}[r]
-0.3280 \\
-0.3390 \\
-0.3900 \\
-0.9460 \\
-0.5840 \\
-1.2790 \\
-0.4520
\end{bmatrix}\text{for scenarios 3 and 4}
\\
\boldsymbol{\zeta} & =
\begin{bmatrix}
-1.0430 &
1.5700 &
0.7720 &
-0.5260
\end{bmatrix}^{\top}
\\
\boldsymbol{\sigma} & =
\begin{bmatrix}
1.1305 &
1.0328 &
1.1673 &
1.2225
\end{bmatrix}^{\top}
\\
\boldsymbol{\Psi} & =
\begin{cases}
\begin{bmatrix}[r]
1.0000 &   -0.2398 &   -0.1834 &    0.2229 \\
-0.2398 &    1.0000 &    0.2550 &   -0.2703\\
-0.1834 &    0.2550 &    1.0000 &   -0.3119\\
0.2229 &   -0.2703 &   -0.3119 &    1.0000
\end{bmatrix} & \text{for scenarios 1 and 3} 
\\
\begin{bmatrix}[r]
1.0000 &   -0.5000 &   -0.5000 &    0.4000 \\
-0.5000 &    1.0000 &    0.4000 &   -0.4000 \\
-0.5000 &    0.4000 &    1.0000 &   -0.4000 \\
0.4000 &   -0.4000 &   -0.4000 &    1.0000 \\
\end{bmatrix} & \text{for scenarios 2 and 4} \\
\end{cases}
\end{align}

\end{appendices}

\end{document}